\documentclass[10pt]{article}

\usepackage{adit_boxed_paper}
\paperpreset{Delta}

\usepackage[round,authoryear]{natbib}
\usepackage{bm}
\usepackage{float}
\usepackage{subcaption}
\usepackage{multirow}
\usepackage{nicefrac}
\usepackage{siunitx}

\newcommand{\NA}{\multicolumn{1}{c}{--}}

\papertitle{Crystalite: A Lightweight Transformer\\for Efficient Crystal Modeling}

\paperauthors{%
Tin Had\v{z}i Veljkovi\'{c}$^{1,2,*}$, Joshua Rosenthal$^{1,2,*}$, Ivor Lon\v{c}ari\'{c}$^{3}$, Jan-Willem van de Meent$^{1,2}$
}

\paperaffiliations{%
$^1$UvA-Bosch Delta Lab, $^2$University of Amsterdam, $^3$Ru\dj er Bo\v{s}kovi\'{c} Institute
}

\papernotes{%
$^*$Equal contribution
}

\paperabstract{%
Generative models for crystalline materials often rely on equivariant graph neural networks, which capture geometric structure well but are costly to train and slow to sample. We present Crystalite, a lightweight diffusion Transformer for crystal modeling built around two simple inductive biases. The first is \emph{Subatomic Tokenization}, a compact atom representation that replaces high-dimensional one-hot encodings and is better suited to continuous diffusion. The second is the \emph{Geometry Enhancement Module} (GEM), which injects periodic minimum-image pair geometry directly into attention through additive geometric biases. Together, these components preserve the simplicity and efficiency of a standard Transformer while making it better matched to the structure of crystalline materials. Crystalite achieves state-of-the-art results on crystal structure prediction benchmarks and strong \textit{de novo} generation performance, attaining the best S.U.N. discovery score among the evaluated baselines while sampling substantially faster than geometry-heavy alternatives.
}

\paperfooter{%
\paperinfoitem{Correspondence:}{THV: \href{mailto:tin.hadzi@gmail.com}{tin.hadzi@gmail.com}; JR: \href{mailto:joshua.rosenthal@student.uva.nl}{joshua.rosenthal@student.uva.nl}}
\paperinfoitem{Code:}{\href{https://github.com/joshrosie/crystalite}{https://github.com/joshrosie/crystalite}}
\paperinfoitem{Keywords:}{Crystal Generation, Crystal Structure Prediction, Diffusion Transformers}
}

\begin{document}
\makepaperheader

\section{Introduction}

The discovery of novel, synthesizable, and diverse crystalline materials with targeted properties remains a central goal of materials science \citep{Merchant2023}. The search space of possible compositions and structures is combinatorially vast, while only a small fraction of candidates is thermodynamically stable. Traditional computational approaches can explore this space systematically \citep{Pickard_2011,Oganov2006}. However, even with large high-throughput infrastructures \citep{jainCommentaryMaterialsProject2013,curtarolo2012aflow,kirklin2015oqmd}, candidate evaluation still typically relies on density functional theory (DFT) \citep{kohnsham, jones2015dft}, which can be computationally expensive \citep{goedecker1999linear}.

Deep generative models offer a promising alternative by learning to propose candidate materials directly from data \citep{xieCrystalDiffusionVariational,zeniGenerativeModelInorganic2025}. In crystal generation, however, the geometric and symmetry structure of the problem has driven much of the literature toward equivariant graph neural networks (GNNs) and other specialized architectures \citep{luoCrystalFlowFlowbasedGenerative2025,jiaoCrystalStructurePrediction2024,zeniGenerativeModelInorganic2025,millerFlowMMGeneratingMaterials2024}. While highly effective, these approaches can be architecturally complex and computationally demanding, motivating the search for simpler approaches that still capture enough crystal geometry to remain competitive \citep{yangScalableDiffusionMaterials2024}. This raises the question: can a lightweight Transformer recover enough geometric structure to compete without explicit equivariant message passing?


Recent work suggests that Transformers can be competitive with GNN-based approaches for crystal generation. In particular, diffusion Transformers have emerged as a promising lightweight alternative for atomistic and crystalline generation \citep{yiCrystalDiTDiffusionTransformer2025a,joshi2025allatom,jinOXtalAllAtomDiffusion2025}. However, these approaches often incorporate crystal geometry only weakly or indirectly, leaving open whether a standard diffusion Transformer can remain simple while benefiting from a more direct injection of periodic geometric structure.

In this work, we introduce \textbf{Crystalite}, a lightweight diffusion Transformer for crystalline materials. Crystalite augments standard multi-head attention with periodic and geometric biases, and uses Subatomic Tokenization in place of high-dimensional one-hot type encodings. This preserves the simplicity and scalability of a standard Transformer backbone while improving its suitability for crystal generation.

Our main contributions are as follows:
\begin{itemize}
    \item We introduce the Geometric Enhancement Module (GEM), a lightweight attention-biasing mechanism that injects periodic and pairwise geometry directly into standard Transformers, providing an efficient alternative to equivariant message passing.
    \item We replace one-hot atom types with Subatomic Tokenization, a compact representation better matched to continuous diffusion.
    \item We show that Crystalite achieves state-of-the-art crystal structure prediction (CSP) and \textit{de novo} generation (DNG) performance, while sampling much faster than geometry-heavy baselines.
    \item We characterize the trade-off between novelty, validity, and stability, and show that MLIP-based stability estimates provide a practical signal for model selection.
\end{itemize}

\begin{figure}[t]
    \centering
    \includegraphics[width=0.9\linewidth]{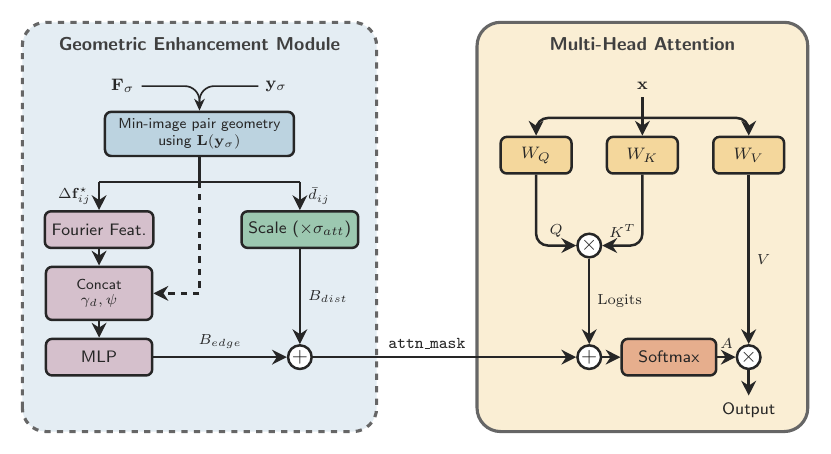}
\caption{
Overview of the proposed architecture. 
\textbf{Left:} The Geometric Enhancement Module (GEM) computes periodic minimum-image pair geometry from $\mathbf F_\sigma$ and the reconstructed lattice $\mathbf L(\mathbf y_\sigma)$.
The resulting geometric features are converted into additive attention biases through a learned edge pathway and a distance-based pathway.
\textbf{Right:} Biases are added to the standard multi-head attention logits before the softmax, allowing attention to depend on crystal geometry while retaining the usual dot-product attention structure.
}
    \label{fig:GEM_figure1}
\end{figure}

\section{Related Work}

Prior work on crystal generation differs largely in how geometric structure is handled. One line of research builds symmetry and periodicity directly into the model through equivariant or geometry-aware architectures. Another explores lighter backbones, including transformers, with weaker inductive bias. Crystalite is most closely related to the recent diffusion-transformer line, but differs in how geometric information is incorporated.

\paragraph{Equivariant and geometry-aware crystal generators.}
Diffusion models \citep{hoDenoisingDiffusionProbabilistic2020,NEURIPS2019_3001ef25} have become a powerful framework for generative modeling in atomistic domains. In crystalline materials, a common strategy is to combine diffusion with equivariant GNNs, since crystal structures naturally admit graph-based representations and are governed by important geometric symmetries (see Appendix \ref{apdx:introduction_to_materials}). MatterGen \citep{zeniGenerativeModelInorganic2025}, for example, is a high-performing equivariant diffusion model built on GemNet \citep{gasteiger_gemnet_2021} that jointly models atom types, fractional coordinates, and lattice parameters, and can also be adapted for inverse design. EGNN \citep{pmlr-v139-satorras21a}, as used in DiffCSP \citep{jiaoCrystalStructurePrediction2024}, has likewise served as the backbone for several subsequent approaches \citep{millerFlowMMGeneratingMaterials2024,hoellmerOpenMaterialsGeneration2025,cornetKineticLangevinDiffusion2025,luoCrystalFlowFlowbasedGenerative2025}. These works also explore increasingly specialized generative formulations to better handle crystal geometry. FlowMM \citep{millerFlowMMGeneratingMaterials2024}, for instance, extends Riemannian flow matching \citep{chenFlowMatchingGeneral2024} to fractional coordinates, while \cite{hoellmerOpenMaterialsGeneration2025} study this setting using stochastic interpolants \citep{JMLR:v26:23-1605}. KLDM \citep{cornetKineticLangevinDiffusion2025} instead handles periodic fractional coordinates by lifting the noising process to an auxiliary flat space using the Lie group structure of the torus. Collectively, these methods show the value of strong geometric inductive bias, but often at the cost of increasing architectural and computational complexity.

\paragraph{Lightweight alternatives to full equivariance.}
A more recent line of work asks whether strong performance in material generation can be achieved without fully equivariant architectures. These approaches are attractive because they are typically simpler, more computationally efficient, and easier to scale. UniMat \citep{yangScalableDiffusionMaterials2024}, for example, shows that a diffusion model based on a 3D U-Net can remain competitive with equivariant baselines and benefit from increased model scale. More broadly, transformer-based approaches have also been explored in autoregressive and hybrid settings, including sequence models over crystal representations \citep{mohantyCrysTextGenerativeAI2024,kazeevWyckoffTransformerGeneration2025,gruverFineTunedLanguageModels2025,caoSpaceGroupInformed2025} and pipelines in which language models provide crystal priors that are later refined by more structured geometric generators \citep{khastagirLLMMeetsDiffusion2025,sriramFlowLLMFlowMatching2024}. These results suggest that fully equivariant message passing may not always be necessary, but they leave open how much geometry a crystal generator should encode directly.

\paragraph{Diffusion transformers for atomistic and crystal generation.}
The works closest to ours are recent diffusion-transformer approaches for molecules, materials, and crystals. ADiT \citep{joshi2025allatom} employs a latent diffusion transformer \citep{peeblesScalableDiffusionModels2023,rombachHighResolutionImageSynthesis2022} with minimal inductive bias for joint generation over molecules and materials, while \cite{moreheadZatom1MultimodalFlow2026} extend this direction with a simpler diffusion-transformer formulation. OXtal \citep{jinOXtalAllAtomDiffusion2025} applies diffusion transformers to crystal structure prediction for metal-organic frameworks and combines this with EDM-style preconditioning and sampling \citep{NEURIPS2022_a98846e9}, while CrystalDiT \citep{yiCrystalDiTDiffusionTransformer2025a} brings the diffusion-transformer type of model to crystalline generation. Crystalite builds most directly on this line of work, but differs in that it injects periodic pairwise geometry directly into attention rather than relying only on augmentation or latent-space structure. In this sense, our goal is not to remove geometric inductive bias, but to incorporate it in a simpler and more modular form than in fully equivariant GNNs.

\begin{figure}[t]
    \centering
    
    \includegraphics[trim=0.1cm 0.1cm 0.0cm 0.0cm, clip, width=0.49\textwidth]{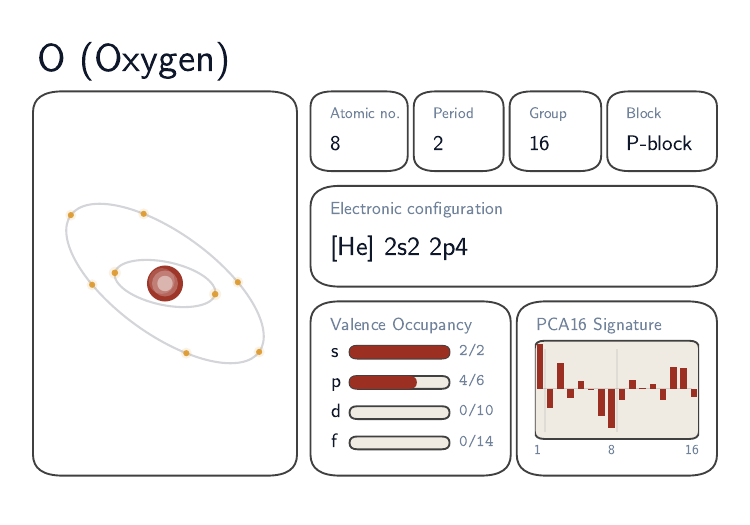}\hfill
    \includegraphics[trim=0.1cm 0.1cm 0.0cm 0.0cm, clip, width=0.49\textwidth]{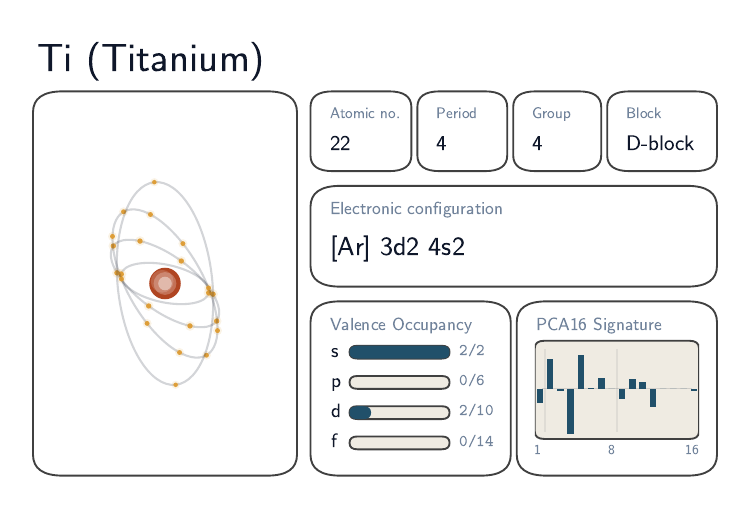}
    
\caption{\textbf{Subatomic tokenization of atomic species.}
Instead of representing each chemical element by a one-hot identity vector, we assign each element a fixed 34-dimensional subatomic descriptor built from its period, group, block, and valence-shell occupancies. These descriptors are compressed to a 16-dimensional token space using PCA, yielding a continuous atom-type representation for diffusion. Diffusion noise is applied in this 16-dimensional space, after which a learned embedding maps the noisy token to the Transformer hidden dimension. Representative examples are shown for oxygen (left) and titanium (right).}
    
    \label{fig:element_comparison}
\end{figure}

\section{Methodology}

Crystalite is built around a simple idea: keep the denoising backbone close to a standard diffusion Transformer, and incorporate crystal-specific structure through the representation, attention mechanism, and sampling procedure. We begin from the standard unit-cell description of a crystal in terms of atom identities, fractional coordinates, and lattice geometry. On top of this representation, we replace one-hot atom identities with subatomic tokens, define diffusion jointly over atom, coordinate, and lattice variables, and process the resulting state with a Transformer that uses one token per atom together with a single global lattice token. Periodic pairwise geometry can then be injected directly into attention through the Geometry Enhancement Module (GEM), while a channel-wise anti-annealing heuristic improves refinement at sampling time.

Concretely, throughout this section we represent a crystal with \(N\) atoms by the unit-cell tuple
\begin{equation}
\mathcal C = (\mathbf A,\mathbf F,\mathbf L), \qquad
\mathbf A\in \{0,1\}^{N\times N_Z},\;
\mathbf F\in [0,1)^{N\times 3},\;
\mathbf L\in \mathbb R^{3\times 3},
\end{equation}
where \(N_Z\) is the number of supported atom types and each row satisfies \(\mathbf A_i=\mathrm{onehot}(a_i)\) for some label \(a_i\in\{1,\ldots,N_Z\}\). Here, \(\mathbf A\) is the atom-type matrix, \(\mathbf F\) contains the fractional coordinates, and \(\mathbf L\) defines the periodic unit cell. The corresponding Cartesian coordinates are given by \(\mathbf X = \mathbf F \mathbf L\).

\subsection{Subatomic Tokenization}
\label{sec:subatomic_tokenization}

A standard representation uses the one-hot atom-type matrix $\mathbf A\in\{0,1\}^{N\times N_Z}$. We found this choice suboptimal for diffusion over crystalline materials for two reasons. First, for realistic materials datasets $N_Z$ can be large (e.g.\ $N_Z=89$ on MP-20), making the atom-type channel unnecessarily high-dimensional relative to the underlying chemical variable. Second, the one-hot geometry is chemically uninformative: all elements are mutually orthogonal, so for example, Li is as far from Na as it is from Xe. This can encourage the model to memorize recurring compositions, while providing no notion of smooth chemical similarity. A learned embedding could in principle reduce dimensionality, but would make the atom-type geometry itself a learned quantity. This adds another representation-learning problem to the denoising objective, and without explicit regularization does not guarantee that chemically similar elements are placed nearby. We therefore use fixed subatomic descriptors, which provide a meaningful continuous geometry from the start of training.

To address this, we replace the one-hot channel by a low-dimensional continuous tokenization, which we refer to as \emph{Subatomic Tokenization}. For each supported element $k\in\{1,\dots,N_Z\}$, let $r_k$, $g_k$, and $b_k$ denote its period, group, and block, and let $(s_k,p_k,d_k,f_k)$ denote its ground-state valence-shell occupancies. The tokenized representation associated with element $k$ is
\begin{equation}
\mathbf h_k
=
\Big[
\mathsf{onehot}(r_k),\;
\mathsf{onehot}(g_k),\;
\mathsf{onehot}(b_k),\;
s_k/2,\; p_k/6,\; d_k/10,\; f_k/14
\Big].
\end{equation}
Figure~\ref{fig:element_comparison} illustrates representative subatomic element tokens. Following the implementation used in our experiments, these element-wise descriptors are standardized across the supported elements, optionally projected with a fixed PCA basis, and finally $\ell_2$-normalized. We continue to denote the resulting tokenized vectors by $\mathbf h_k$. The tokenized atom matrix is then
\begin{equation}
\mathbf{H} = \big[ \mathbf{h}_{a_1}, \dots, \mathbf{h}_{a_N} \big]^{\top} \in \mathbb{R}^{N \times d_H},
\end{equation}
where $d_H$ denotes the token dimension after optional PCA compression. This design serves two purposes. First, it reduces the dimensionality of the atom-type channel, which makes denoising statistically easier and lowers the capacity of the model to memorize frequent compositional patterns. Second, it equips the diffusion process with a chemically meaningful geometry: errors in subatomic space become structured, so that under noise the model is encouraged to confuse elements with plausible substitutions before unrelated species.

Subatomic Tokenization is especially natural in our EDM formulation, since atom types are treated as continuous diffusion variables jointly with fractional coordinates and lattice parameters. The denoiser therefore does not need to recover a sparse one-hot vector in a high-dimensional simplex-like space, but instead returns a low-dimensional subatomic token. During sampling, the denoised token $\hat{\mathbf h}_i$ is mapped back to a discrete element by nearest-token decoding,
\begin{equation}
\hat a_i
=
\arg\max_{k \in \{1,\dots,N_Z\}}
\langle \hat{\mathbf h}_i, \mathbf h_k \rangle,
\end{equation}
which is equivalent to cosine-similarity decoding because all token vectors are normalized. This keeps the training and decoding geometries aligned. In the crystal structure prediction (CSP) setting, where the composition is known, the atom-type features are held fixed and only the coordinate and lattice channels are denoised. We provide additional information on this embedding in Appendix \ref{apdx:subatomic_tokenization_details}, and ablate it against one-hot element encodings in Appendix~\ref{apdx:ablation_subatomic}.

\subsection{Diffusion formulation for crystals}

Starting from a crystal \(\mathcal C=(\mathbf A,\mathbf F,\mathbf L)\), we define a continuous diffusion state
\[
(\mathbf H,\mathbf F,\mathbf y),
\]
where \(\mathbf H\) is the subatomic atom-type representation, \(\mathbf F\in[0,1)^{N\times 3}\) contains the fractional coordinates, and \(\mathbf y\in\mathbb R^6\) is a latent parameterization of the lattice. Concretely, \(\mathbf h_i\in\mathbb R^{d_H}\) denotes the token of atom \(i\), and \(\mathbf H=[\mathbf h_1,\dots,\mathbf h_N]^\top\). Likewise, \(\mathbf f_i\in[0,1)^3\) denotes the fractional coordinate of atom \(i\), and \(\mathbf F=[\mathbf f_1,\dots,\mathbf f_N]^\top\).

Rather than diffusing the raw lattice matrix \(\mathbf L\in\mathbb R^{3\times 3}\) directly, we represent it through a lower-triangular latent \(\mathbf y\in\mathbb R^6\) and reconstruct
\begin{equation}
\mathbf L(\mathbf y)=
\begin{bmatrix}
e^{y_1} & 0 & 0\\
y_2 & e^{y_3} & 0\\
y_4 & y_5 & e^{y_6}
\end{bmatrix}.
\label{eq:lattice_param}
\end{equation}
This yields a stable unconstrained representation with positive diagonal entries and reduces representational redundancy in the lattice channel. The diffusion model therefore operates on the continuous tuple \((\mathbf H,\mathbf F,\mathbf y)\).

The lattice representation remains basis-dependent, however. To reduce basis ambiguity, we preprocess each structure into a Niggli-reduced cell and express the lattice in a fixed lattice-parameter convention before tokenization. During training, the only explicit crystal augmentation is a random global translation of the fractional coordinates; we do not augment over lattice-basis permutations or other equivalent cell choices.

We adopt the EDM framework because it provides a well-optimized set of preconditioning, loss weighting, and sampling choices. Following EDM, at each training step we sample a noise level from
\[
\log \sigma \sim \mathcal N(P_{\mathrm{mean}},P_{\mathrm{std}}^2),
\]
and perturb all three channels jointly:
\begin{equation}
(\mathbf H_\sigma,\mathbf F_\sigma,\mathbf y_\sigma)
=
(\mathbf H,\mathbf F,\mathbf y)+\sigma\,\bm{\varepsilon},
\label{eq:joint_noising}
\end{equation}
where \(\bm{\varepsilon}\) denotes Gaussian noise with the appropriate channel-wise shapes. For the coordinate channel, noise is added in a centered Euclidean representation: fractional coordinates are first shifted to a centered cube, Gaussian noise is added in that space, and the resulting noisy coordinates are wrapped back into \([0,1)^3\) before being embedded by the Transformer. The training loss, however, is evaluated using a componentwise wrapped residual in fractional space. This respects periodicity on the torus, but unlike GEM it is not a metric-aware minimum-image search under the lattice metric. Full details are given in Appendix~\ref{apdx:edm}. As in EDM, the noisy inputs and raw network outputs are combined through the standard channel-wise preconditioning coefficients \(c_{\mathrm{in}}(\sigma)\), \(c_{\mathrm{skip}}(\sigma)\), and \(c_{\mathrm{out}}(\sigma)\); we defer the exact formulas to Appendix~\ref{apdx:edm}.

We train the model with separate denoising losses for the atom-type, coordinate, and lattice channels. Atom tokens and lattice latents are regressed directly in Euclidean space, while coordinates are compared through componentwise wrapped residuals in fractional space. Writing \(\mathrm{wrap}(\mathbf u)=\mathbf u-\mathrm{round}(\mathbf u)\), the three channel-wise losses are
\begin{equation}
\mathcal L_H=\frac{1}{N}\sum_{i=1}^N w_H(\sigma)\,\|\hat{\mathbf h}_i-\mathbf h_i\|_2^2,
\qquad
\mathcal L_F=\frac{1}{N}\sum_{i=1}^N w_F(\sigma)\,\big\|\mathrm{wrap}(\hat{\mathbf f}_i-\mathbf f_i)\big\|_2^2,
\qquad
\mathcal L_{\mathrm{lat}}=\frac{1}{6}\,w_{\mathrm{lat}}(\sigma)\,\|\hat{\mathbf y}-\mathbf y\|_2^2.
\end{equation}
The total objective is
\begin{equation}
\mathcal L
=
\lambda_H \mathcal L_H
+
\lambda_F \mathcal L_F
+
\lambda_{\mathrm{lat}} \mathcal L_{\mathrm{lat}},
\end{equation}
where \(w_H(\sigma)\), \(w_F(\sigma)\), and \(w_{\mathrm{lat}}(\sigma)\) are the standard EDM channel-wise weights.

We use the same diffusion formulation for both \textit{de novo} generation and crystal structure prediction. In DNG, Crystalite models the joint distribution \(p_\theta(\mathbf A,\mathbf F,\mathbf L)\) and generates all channels jointly. Because the number of atoms per unit cell varies across structures, we first sample \(N \sim p(N)\) from the empirical training-set distribution and then generate the atom-type, coordinate, and lattice channels for that sampled size. In CSP, it instead models the conditional distribution \(p_\theta(\mathbf F,\mathbf L\mid \mathbf A)\), treating structure prediction as conditional generation with the composition fixed.

\subsection{Crystalite architecture}

\begin{figure}[t]
    \centering
    \includegraphics[width=0.95\linewidth]{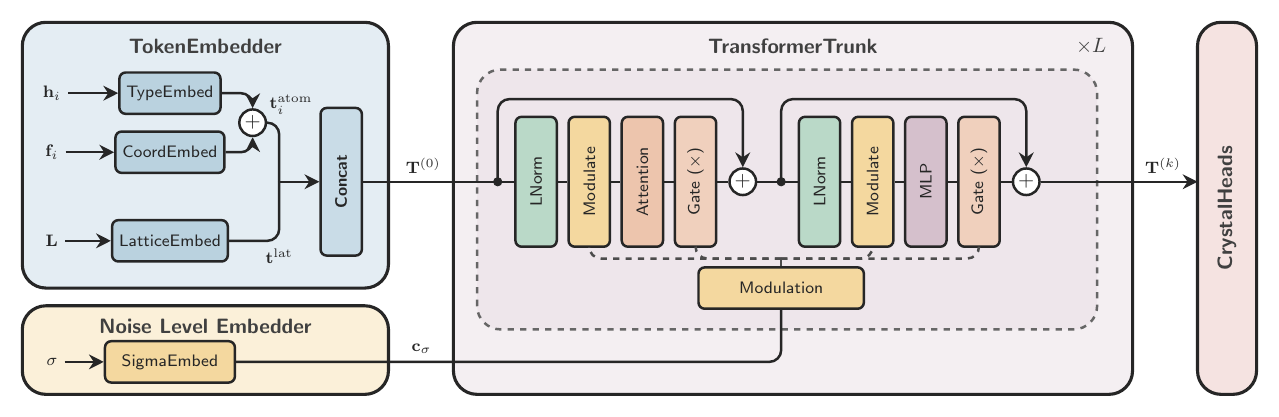}
    \caption{
    Overview of the Crystalite architecture. The model operates on the continuous crystal state \((\mathbf H,\mathbf F,\mathbf y)\). Atom-type and coordinate embeddings are added to form one token per atom, while the lattice embedding produces a single global lattice token. The resulting sequence is processed by an AdaLN-conditioned Transformer trunk, and output heads predict \(\hat{\mathbf H}\), \(\hat{\mathbf F}\), and \(\hat{\mathbf y}\).
    }
    \label{fig:crystalite_architecture}
\end{figure}

Crystalite operates on the continuous diffusion state \((\mathbf H,\mathbf F,\mathbf y)\) using a standard Transformer backbone with one token per atom and one additional token for the lattice. The full Crystalite architecture is shown in Figure \ref{fig:crystalite_architecture}.

\paragraph{Input parameterization.}
For each atom \(i\), we map the subatomic token \(\mathbf h_i \in \mathbb R^{d_H}\) and the corresponding fractional coordinate \(\mathbf f_i \in \mathbb R^3\) into a common hidden dimension through separate learned embedders. These are then added to form a single atom token,
\begin{equation}
\mathbf t_i^{\mathrm{atom}} = E_H(\mathbf h_i) + E_F(\mathbf f_i),
\end{equation}
where \(E_H\) and \(E_F\) denote the atom-type and coordinate embedders. In this way, each atom token jointly represents chemical identity and geometric position. The lattice is embedded separately. The latent lattice vector \(\mathbf y \in \mathbb R^6\) is mapped to a single global lattice token,
\begin{equation}
\mathbf t^{\mathrm{lat}} = E_{\mathrm{lat}}(\mathbf y).
\end{equation}
For a crystal with \(N\) atoms, the full input sequence is therefore
\begin{equation}
\mathbf T^{(0)}
=
\big[
\mathbf t_1^{\mathrm{atom}},\dots,\mathbf t_N^{\mathrm{atom}},\mathbf t^{\mathrm{lat}}
\big]
\in \mathbb R^{(N+1)\times d},
\end{equation}
where \(d\) is the model width. The diffusion noise level is embedded through a small MLP applied to the standard EDM noise coordinate \(c_{\mathrm{noise}}(\sigma)=\tfrac{1}{4}\log\sigma\), producing a conditioning vector \(\mathbf c_\sigma \in \mathbb R^d\) that is injected into every block through adaptive layer normalization (AdaLN).

\paragraph{Output parameterization.}
The sequence \(\mathbf T^{(0)}\) is processed by a standard Transformer backbone composed of stacked self-attention and feed-forward blocks. We denote the state after $K$ layers as:
\[
\mathbf T^{(K)}
=
\big[
\mathbf t_1^{(K)},\dots,\mathbf t_N^{(K)},\mathbf t_{\mathrm{lat}}^{(K)}
\big]
\]
The first \(N\) tokens are then decoded into denoised atom-token and coordinate predictions, while the final token is decoded into the lattice latent:
\begin{equation}
\hat{\mathbf h}_i = D_H(\mathbf t_i^{(K)}),
\qquad
\hat{\mathbf f}_i = D_F(\mathbf t_i^{(K)}),
\qquad
\hat{\mathbf y} = D_{\mathrm{lat}}(\mathbf t_{\mathrm{lat}}^{(K)}).
\end{equation}
Collecting these predictions over all atoms gives
\begin{equation}
(\hat{\mathbf H},\hat{\mathbf F},\hat{\mathbf y})
=
\mathrm{Crystalite}_{\theta}(\mathbf H_\sigma,\mathbf F_\sigma,\mathbf y_\sigma;\sigma),
\end{equation}
which are interpreted as denoised predictions and combined with the noisy inputs through the EDM preconditioning rules described in Appendix~\ref{apdx:edm}. A more detailed architectural description is provided in Appendix~\ref{apdx:architecture}.

\subsection{Geometry Enhancement Module (GEM)}
\label{sec:gem}

Crystalite augments standard self-attention with a geometry-dependent additive bias, recomputed at each denoising step. 
This design is related in spirit to additive structural biases used in graph transformers such as Graphormer \citep{graphormer}, but here the bias is constructed from periodic minimum-image crystal geometry.
This injects periodic pairwise structure into the attention mechanism without requiring equivariant message-passing, as shown in Figure \ref{fig:GEM_figure1}.

Given the fractional coordinates $\mathbf{F}$ and lattice latent $\mathbf{y}$, we reconstruct the lattice matrix $\mathbf{L}(\mathbf{y})$. For each atom pair $(i,j)$, we compute the minimum-image fractional displacement $\Delta\mathbf{f}_{ij}^{\star}$ under periodic boundary conditions and its normalized Cartesian distance:
\begin{equation}
    \bar{d}_{ij} = \frac{\|\Delta\mathbf{f}_{ij}^{\star}\mathbf{L}(\mathbf{y})\|_2}{s(\mathbf{y})},
\end{equation}
where \(s(y)\) is a characteristic cell scale; in our implementation we use the mean of the three lattice lengths. Unlike the wrapped fractional residual used in the coordinate loss, GEM selects the periodic image by minimizing the Cartesian quadratic form induced by the lattice metric \(\mathbf G=\mathbf L\mathbf L^\top\).

From this geometry, GEM constructs a head-wise attention bias by combining a direct distance penalty with learned edge features. This combined bias is then modulated by a learned noise-dependent gate $g_h(\sigma)$ to form the final geometric bias:
\begin{equation}
    B^{\mathrm{geom}}_{hij} = g_h(\sigma) \left( B^{\mathrm{dist}}_{hij} + B^{\mathrm{edge}}_{hij} \right)
\end{equation}
where the distance penalty $B^{\mathrm{dist}}_{hij} = w_h \bar{d}_{ij}$ uses a learned, monotonically non-positive slope $w_h \le 0$, and the edge bias models non-linear interactions through an MLP:
\begin{equation}
    B^{\mathrm{edge}}_{hij} = \operatorname{MLP}_{\mathrm{edge}}\!\left(\left[\gamma_{\Delta}(\Delta\mathbf{f}_{ij}^{\star}), \gamma_d(\bar{d}_{ij}), \psi(\mathbf{y})\right]\right)_h.
\end{equation}
Here, $\gamma_{\Delta}$ applies Fourier features to the displacement, $\gamma_d$ applies a Radial Basis Function (RBF) kernel to the distance, and $\psi(\mathbf{y})$ is a low-dimensional lattice descriptor. 

This geometric bias is applied exclusively to atom--atom interactions. Padding $\mathbf{B}^{\mathrm{geom}}$ with zeros for any interactions involving the global lattice token, the attention update becomes:
\begin{equation}
    \operatorname{Attn}(Q,K,V) = \operatorname{softmax}\!\left(\frac{QK^\top}{\sqrt{d}} + \mathbf{B}^{\mathrm{geom}}\right)V.
\end{equation}
This allows the model to emphasize geometrically compatible atom pairs directly in the attention logits while maintaining the simplicity and efficiency of a standard diffusion Transformer. We provide more details on the implementation in Appendix \ref{apdx:GEM}.

\subsection{Channel-wise anti-annealing during sampling.}
During EDM sampling, we optionally apply a \emph{channel-wise anti-annealing} step, which rescales the reverse-time update separately for the atom-token, coordinate, and lattice channels. Intuitively, this acts as a channel-dependent time warp: if a particular channel denoises more slowly or dominates the remaining error, anti-annealing drives that channel more aggressively toward the denoised prediction while leaving the learned denoiser itself unchanged. This was particularly useful in our setting for improving geometric refinement at sampling time without modifying the training objective. Concretely, for each channel \(q\in\{H,F,\mathrm{lat}\}\), we replace the standard Heun-style EDM update by
\begin{equation}
\mathbf z_{i+1}^{(q)}
=
\bar{\mathbf z}_i^{(q)}
+
(\sigma_{i+1}-\bar{\sigma}_i)\,
\alpha_i^{(q)}\,
\frac{
\mathbf d_i^{(q)}+\mathbf d_{i+1}^{(q),E}
}{2},
\qquad
\alpha_i^{(q)} \geq 1,
\end{equation}
where \(\alpha_i^{(q)}\) is a channel-specific anti-annealing factor derived from an auxiliary Karras schedule, and \(\alpha_i^{(q)}=1\) recovers the standard EDM sampler. Full details are given in Appendix~\ref{apdx:anti_annealing}, with sensitivity studies for DNG and CSP in Appendices~\ref{apdx:dng_anti_annealing} and~\ref{apdx:csp_anti_annealing}.

\begin{figure}[t]
    \centering
    \includegraphics[width=1.0\linewidth]{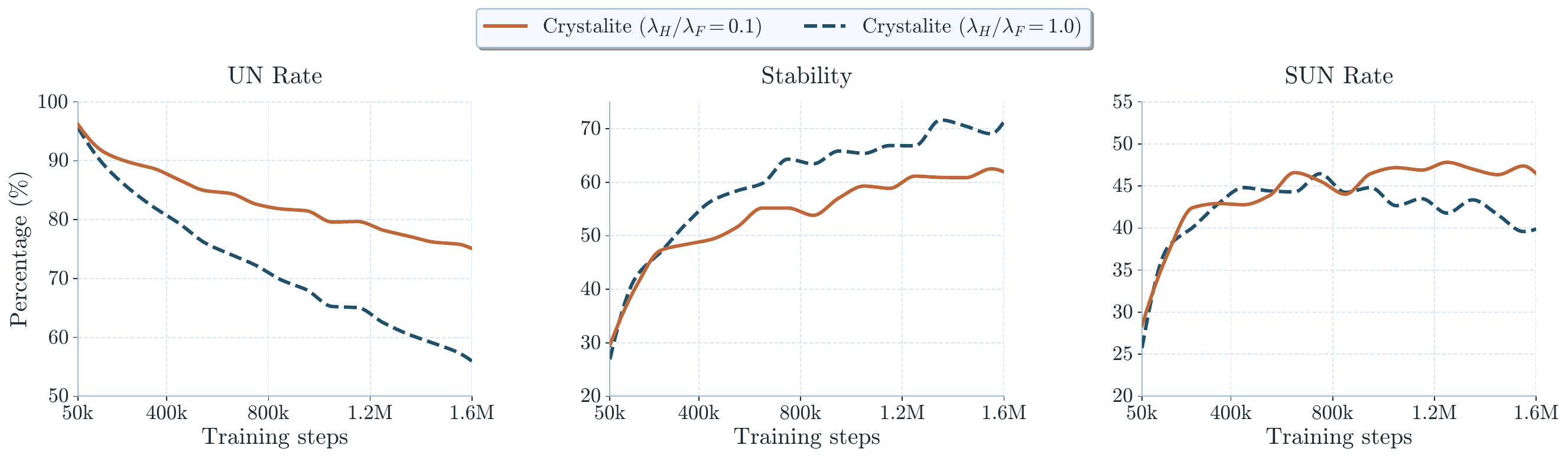}
\caption{\textbf{Training-time trade-off in \textit{de novo} generation.} UN rate (left), stability (middle), and SUN rate (right) as a function of training steps for two Crystalite runs with different atom-loss settings. The setting that achieves higher stability also loses UN more quickly, whereas the more diversity-preserving setting yields a flatter and more sustained SUN trajectory. Overall, the figure illustrates the central DNG trade-off: improved distributional fit tends to increase stability, but often at the cost of novelty and uniqueness, making checkpoint selection and loss balancing important in practice.}
    \label{fig:training_tradeoff}
\end{figure}

\section{Experimental Setup}
\subsection{Datasets}
We use three realistic datasets to benchmark the models: {MP-20} \citep{xieCrystalDiffusionVariational}, a subset of the Materials Project \citep{jainCommentaryMaterialsProject2013} containing 45 231 crystalline materials of up to 20 atoms per unit cell with 89 distinct atom types; {MPTS-52} \citep{Baird2024} where each split is derived chronologically from the Materials Project and contains 40 476 structures with up to 50 atoms per unit cell -- notably the temporal component adds an extra degree of difficulty where the training, validation, and test sets exhibit a fundamental shift in their underlying distributions, making this benchmark particularly challenging; and {Alex-MP-20} \citep{zeniGenerativeModelInorganic2025} which contains 675 204 structures with up to 20 atoms per unit cell, derived from Alexandria and MP-20. Here we follow the data splits as given by \cite{hoellmerOpenMaterialsGeneration2025}.

\subsection{Task setup}
We evaluate Crystalite in two settings: \textit{de novo} generation (DNG) and crystal structure prediction (CSP). In the DNG setting, the model generates atom types, fractional coordinates, and lattice parameters jointly from noise. In the CSP setting, the atomic composition is provided as input, and the model predicts only the crystal geometry, i.e.\ the fractional coordinates and lattice. Operationally, this is implemented by fixing the atom-type features to the known composition and masking the type loss during training and sampling.

\paragraph{Model settings.} We use the same Crystalite model family for DNG and CSP, but the reported experiments use task-specific configurations. The DNG model uses a \(14\)-layer Transformer with width \(d=512\) and PCA-compressed Subatomic Tokenization, while the CSP models use the same depth with a wider \(d=1024\) Transformer and fixed atomic-number type features. The Alex-MP-20 CSP experiment uses the same model configuration as MP-20, changing only the dataset. All main configurations use \(16\) attention heads, \texttt{bfloat16} training, EMA weights for sampling and evaluation, and GEM, with task- and dataset-specific distance- and edge-bias settings. Full architectural, training, and sampling hyperparameters are provided in Table~\ref{tab:crystalite_base_config}; additional EDM details are provided in Appendix~\ref{apdx:edm}.

\paragraph{Sampling speed benchmarking.}
For a fair comparison of sampling speed, we measure the wall-clock time required to generate 1,000 crystals on a single NVIDIA H100 GPU. For each model, we use the largest sampling batch size that fits in memory, so that each method is evaluated at its highest feasible throughput. Unless otherwise noted, the reported timing corresponds to the standard inference setting used for cross-model comparison. For Crystalite, we additionally report a second timing, marked with $^{\dag}$ in Table~\ref{tab:gen_quality}, obtained with optimized bfloat16 inference. We regard the primary timing as the main comparison across methods, and the daggered number as a reference for the throughput attainable by Crystalite under an optimized implementation.


\section{Results and Discussion}

\subsection{CSP Results}
Table~\ref{tab:main_results} summarizes the results on the CSP benchmarks. Across all datasets, Crystalite outperforms prior methods. Using Match Rate to assess successful structure recovery and RMSE to measure geometric accuracy (see Appendix~\ref{apdx:evaluation_metrics}), Crystalite achieves state-of-the-art results on both criteria. The improvement is especially pronounced in RMSE, indicating more accurate structural recovery even in settings where match-based performance is already strong.

The effect of GEM is examined in more detail in Appendix~\ref{apdx:GEM_CSP}. Compared to the best previous RMSE on each benchmark, Crystalite reduces RMSE by 34.8\%, 44.3\%, and 74.7\% on MP-20, MPTS-52, and Alex-MP-20, respectively, for an average reduction of approximately 51\%. The GEM ablation shows that this gain is partly due to the geometry-aware attention bias: GEM has little effect on Match Rate, but consistently improves RMSE by about 20\%. This suggests that GEM mainly refines geometric accuracy rather than determining whether the correct structural mode is recovered.

 \begin{table}[t]
\centering
\small
\caption{Crystal structure prediction results across standard benchmarks. Best values are in bold.}
\label{tab:main_results}
\renewcommand{\arraystretch}{1.1}
\begin{tabular}{@{} l *{3}{S[table-format=2.2] S[table-format=1.4]} @{}}
\toprule
\multirow{3}{*}{\textbf{Model}} &
\multicolumn{2}{c}{\textbf{MP-20}} &
\multicolumn{2}{c}{\textbf{MPTS-52}} &
\multicolumn{2}{c}{\textbf{Alex-MP-20}} \\
\cmidrule(lr){2-3}\cmidrule(lr){4-5}\cmidrule(lr){6-7}
& {MR} & {RMSE} & {MR} & {RMSE} & {MR} & {RMSE} \\
& {(\%) $\uparrow$} & {$\downarrow$}
& {(\%) $\uparrow$} & {$\downarrow$}
& {(\%) $\uparrow$} & {$\downarrow$} \\
\midrule
CDVAE           & 33.90 & 0.1045 &  5.34 & 0.2106 & \NA & \NA \\
DiffCSP         & 51.49 & 0.0631 & 12.19 & 0.1786 & \NA & \NA \\
FlowMM          & 61.39 & 0.0566 & 17.54 & 0.1726 & \NA & \NA \\
CrystalFlow     & 62.02 & 0.0710 & 22.71 & 0.1548 & \NA & \NA \\
KLDM            & 65.83 & 0.0517 & 23.93 & 0.1276 & \NA & \NA \\
OMatG           & 63.75 & 0.0720 & 25.15 & 0.1931 & 64.71 & 0.1251 \\
\midrule
\textbf{Crystalite} &
{\bfseries 66.09} & {\bfseries 0.0337} & {\bfseries 31.56} & {\bfseries 0.0711} & {\bfseries 68.26} & {\bfseries 0.0317} \\
& {\scriptsize $\pm$ 0.09} & {\scriptsize $\pm$ 0.005} & {\scriptsize $\pm$ 0.06} & {\scriptsize $\pm$ 0.0010} & {\scriptsize $\pm$ 0.12} & {\scriptsize $\pm$ 0.0004} \\
\bottomrule
\end{tabular}
\end{table}

\subsection{DNG Results}
Table~\ref{tab:gen_quality} summarizes the main \textit{de novo} generation results. Crystalite achieves the highest SUN rate and the fastest sampling speed among the compared methods. Since \textit{de novo} generation is fundamentally governed by a trade-off between stability and diversity, we treat SUN as the primary summary metric. The remaining reported metrics can be grouped into two broad categories: \emph{quality and diversity} metrics, and \emph{stability and distribution} metrics, which are described in detail in Appendix~\ref{apdx:evaluation_metrics}. In practice, however, these quantities are tightly coupled, so model selection depends strongly on which aspect of performance is prioritized. As shown in Figure~\ref{fig:training_tradeoff}, training induces a clear trade-off. As optimization progresses, the model more closely matches the training distribution, which tends to improve validity, stability, and distributional alignment, but at the same time reduces novelty and uniqueness. Since the ground-truth dataset itself lacks perfect compositional validity, very high scores on these metrics typically signal overfitting and lower novelty rates.

This trade-off is especially pronounced because atom types are modeled jointly with coordinates and lattice parameters, making compositional memorization difficult to control independently of structural quality. We mitigate this by substantially downweighting the atom-type loss, which Figure~\ref{fig:training_tradeoff} shows yields a flatter, more sustained SUN trajectory and more stable checkpoint selection compared to evenly balanced loss weights. As further analyzed in the GEM ablation study (Appendix~\ref{apdx:GEM_DNG}), GEM mainly improves the stability side of this trade-off, contributing to consistently higher SUN throughout training.

\begin{table}[t]
    \centering
    \small
\caption{Generative quality, diversity, stability, distribution, and sampling speed metrics. All metrics are computed from 10{,}000 generated crystals per model. Stability-based quantities are evaluated using the same NequIP-based relaxation pipeline for all methods. Sampling time is reported in seconds per 1k generated crystals; for Crystalite, $^{\dag}$ denotes an optimized implementation.}
    \label{tab:gen_quality}
    \renewcommand{\arraystretch}{1.1}
    \setlength{\tabcolsep}{4pt}
    \begin{tabular}{@{} l *{5}{S[table-format=3.2]} S[table-format=2.2] S[table-format=2.2] S[table-format=1.3] S[table-format=1.3] S[table-format=4.1] @{}}
        \toprule
        \multirow{3}{*}{\textbf{Model}}
        & \multicolumn{5}{c}{\textbf{Quality and Diversity}}
        & \multicolumn{5}{c}{\textbf{Stability, Distribution, and Speed}} \\
        \cmidrule(lr){2-6} \cmidrule(lr){7-11}
        & {Struct. Val.}
        & {Comp. Val.}
        & {Unique}
        & {Novel}
        & {U.N.}
        & {Stable}
        & {S.U.N.}
        & {wdist-$\rho$}
        & {wdist N-ary}
        & {Time/1k} \\
        & {(\%) $\uparrow$}
        & {(\%) $\uparrow$}
        & {(\%) $\uparrow$}
        & {(\%) $\uparrow$}
        & {(\%) $\uparrow$}
        & {(\%) $\uparrow$}
        & {(\%) $\uparrow$}
        & {$\downarrow$}
        & {$\downarrow$}
        & {(s) $\downarrow$} \\
        \midrule
        FlowMM & 93.03 & 83.15 & 97.44 & 85.00 & 83.99 & 46.05 & 31.64 & 1.389 & {\bfseries 0.075} & {1560} \\
        CrystalDiT & 77.82 & 67.28 & 90.88 & 59.33 & 56.86 & {\bfseries 83.41} & 41.70 & 0.202 & 0.171 & {73.72} \\
        DiffCSP & {\bfseries 99.93} & 82.10 & 96.90 & 89.53 & 87.89 & 50.28 & 38.60 & 0.192 & 0.344 & {237} \\
        MatterGen & 99.78 & 83.72 & {\bfseries 98.10} & {\bfseries 91.14} & {\bfseries 90.26} & 51.70 & 42.29 & 0.088 & 0.184 & {2639} \\
        ADiT & 99.52 & {\bfseries 90.15} & 90.25 & 59.80 & 56.91 & 76.90 & 36.76 & 0.231 & 0.089 & {84.81} \\
        \midrule
        \textbf{Crystalite} & 99.61 & 81.72 & 95.24 & 79.30 & 77.33 & 69.72 & {\bfseries 47.49} & {\bfseries 0.051} & 0.127 & {\textbf{22.36/5.14$^{\dag}$}}  \\
        & {\scriptsize $\pm$ 0.06} & {\scriptsize $\pm$ 0.24} & {\scriptsize $\pm$ 0.19} & {\scriptsize $\pm$ 0.12} & {\scriptsize $\pm$ 0.21} & {\scriptsize $\pm$ 0.85} & {\scriptsize $\pm$ 0.77} & {\scriptsize $\pm$ 0.010} & {\scriptsize $\pm$ 0.006} & \\
        \bottomrule
    \end{tabular}
\end{table}

\paragraph{Fairness and comparability between models.}
Our primary evaluation pipeline uses NequIP-based relaxation \citep{nequip} together with SUN-based checkpoint selection. For fairness, all baseline results reported in the main tables were obtained by evaluating the competing methods within this same pipeline, rather than by taking published numbers at face value. Nevertheless, since those methods may originally have been trained and checkpointed under different criteria, it remains important to verify that Crystalite does not benefit disproportionately from our setup. We therefore also evaluate Crystalite under external benchmarking pipelines, namely the MatterGen \citep{zeniGenerativeModelInorganic2025} evaluation pipeline and LeMat GenBench \citep{betala2026lematgenbenchunifiedevaluationframework}; the corresponding results are reported in Table~\ref{tab:lemat-genbench} and Appendix Table~\ref{tab:mattergen_pipeline}.

\paragraph{Extensive and intensive metrics.}
In \textit{de novo} generation, evaluation metrics do not all behave the same way as the number of generated samples increases. Some reflect properties of an individual draw and can therefore be estimated reliably from random subsets. Others instead characterize the generated set as a whole and vary systematically with the total sampling budget. By analogy with physics, we refer to these as sample-\emph{intensive} and sample-\emph{extensive} metrics, respectively. Uniqueness, and derived quantities such as the UN rate, are strongly sample-extensive: as more crystals are generated, duplicates inevitably accumulate, so these metrics typically decrease.

\begin{figure}[t]
    \centering
    \includegraphics[width=1.0\linewidth]{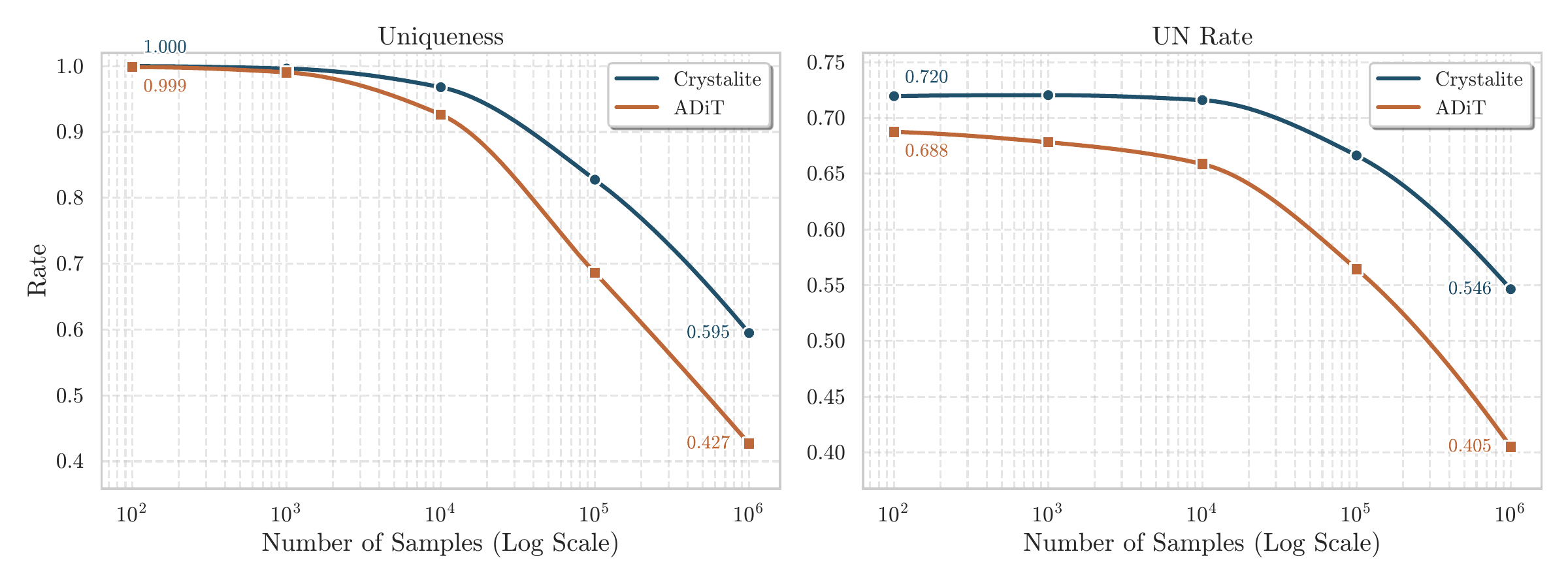}
\caption{\textbf{Large-scale generation.} Uniqueness and unique-and-novel (UN) rate are shown as a function of the number of generated crystals for Crystalite and ADiT. Crystalite consistently preserves more diversity at scale, reaching a higher UN rate at $10^6$ samples and higher uniqueness.}
    \label{fig:large_scale}
\end{figure}

This dependence matters in practice, since a useful crystal generator should not only produce plausible structures, but should also continue to discover many distinct and previously unseen candidates at scale. We therefore compare Crystalite and ADiT in Figure~\ref{fig:large_scale} as a function of the number of generated crystals, showing that Crystalite preserves diversity more effectively as sampling is scaled up. More broadly, this suggests that sample-extensive metrics should always be reported together with the total number of generated samples, since their values are not directly comparable across different budgets. We discuss this issue further in Appendix~\ref{apdx:intensive_extensive_metrics}, where we formalize the distinction and clarify which metrics can, and cannot, be reliably estimated from subsets.

\begin{table}[H]
    \centering
    \small
    \caption{Generation, stability, and relaxation metrics for MP-20 trained models on the LeMat-GenBench leaderboard \citep{betala2026lematgenbenchunifiedevaluationframework}, separated by relaxation status.}
    \label{tab:lemat-genbench}
    \renewcommand{\arraystretch}{1.1}
    \setlength{\tabcolsep}{4pt}
    \begin{tabular}{@{} l *{7}{S[table-format=2.2]} S[table-format=1.4] S[table-format=1.4] @{}}
        \toprule
        \multirow{2}{*}{\textbf{Model}} 
        & {Valid} 
        & {Unique} 
        & {Novel} 
        & {Stable} 
        & {Metastable} 
        & {SUN} 
        & {MSUN} 
        & {E Above Hull} 
        & {Relax. RMSD} \\
        & {(\%) $\uparrow$} 
        & {(\%) $\uparrow$} 
        & {(\%) $\uparrow$} 
        & {(\%) $\uparrow$} 
        & {(\%) $\uparrow$} 
        & {(\%) $\uparrow$} 
        & {(\%) $\uparrow$} 
        & {(eV) $\downarrow$} 
        & {(\AA) $\downarrow$} \\
        \midrule
        \multicolumn{10}{c}{\textbf{Pre-Relaxed Models}} \\
        \midrule
        WyFormer [\citenum{kazeevWyckoffTransformerGeneration2025}] & 93.40 & 93.00 & 66.40 & 0.50 & 15.70 & 0.10 & 1.90 & 0.4988 & 0.8121 \\
        WyFormer-DFT [\citenum{kazeevWyckoffTransformerGeneration2025}] & 95.20 & 95.00 & 66.40 & 3.70 & 24.80 & 0.40 & 7.80 & 0.2708 & 0.4173 \\
        PLaID++ [\citenum{xu2025plaidpreferencealignedlanguage}] & 96.00 & 77.80 & 24.20 & 12.40 & {\bfseries 60.70} & 1.00 & 7.60 & {\bfseries 0.0854} & 0.1286 \\
        MatterGen [\citenum{zeniGenerativeModelInorganic2025}] & 95.70 & 95.10 & {\bfseries 70.50} & 2.00 & 33.40 & 0.20 & 15.00 & 0.1834 & 0.3878 \\
        OMatG [\citenum{hoellmerOpenMaterialsGeneration2025}] & 96.40 & 95.20 & 51.20 & 11.60 & 49.80 & 1.00 & 18.00 & 0.0956 & {\bfseries 0.0759} \\
        \textbf{Crystalite} & {\bfseries 97.20} & {\bfseries 95.80} & 53.20 & {\bfseries 12.70} &   51.60 & {\bfseries 1.50} & {\bfseries 22.60} & 0.0905 & 0.1320 \\
        \midrule
        \multicolumn{10}{c}{\textbf{Non-Pre-Relaxed Models}} \\
        \midrule
        Crystal-GFN [\citenum{mistal2023crystalgfn}] & 51.70 & 51.70 & 51.70 & 0.00 & 0.00 & 0.00 & 0.00 & 2.0858 & 1.8665 \\
        ADiT [\citenum{joshi2025allatom}] & 90.60 & 87.80 & 26.00 & 0.40 & {\bfseries 36.50} & 0.00 & 1.00 & 0.3333 & 0.3794 \\
        CrystalFormer [\citenum{caoSpaceGroupInformed2025}] & 69.90 & 69.40 & 31.80 & 1.40 & 28.80 & 0.00 & 3.10 & 0.7039 & 0.6585 \\
        SymmCD [\citenum{levy2025symmcd}] & 73.40 & 73.00 & 47.00 & 1.40 & 18.60 & 0.10 & 2.40 & 0.8761 & 0.8720 \\
        DiffCSP++ [\citenum{jiao2024space}] & 95.30 & {\bfseries 95.10} & 62.00 & 1.00 & 26.40 & {\bfseries 0.20} & 5.00 & 0.4093 & 0.6933 \\
        DiffCSP [\citenum{jiaoCrystalStructurePrediction2024}] & {\bfseries 95.70} & 94.80 & {\bfseries 66.20} & {\bfseries 2.30} & 29.80 & 0.10 & {\bfseries 8.50} & {\bfseries 0.2747} & {\bfseries 0.3794} \\
        \bottomrule
    \end{tabular}
\end{table}

\section{Conclusion}

We introduced Crystalite, a lightweight diffusion Transformer for crystal structure prediction and \textit{de novo} crystal generation. By combining Subatomic Tokenization with the Geometry Enhancement Module (GEM), Crystalite injects crystal-specific inductive bias into a standard Transformer without relying on expensive equivariant message passing.

Across benchmarks, Crystalite achieves state-of-the-art crystal structure prediction performance and strong \textit{de novo} generation results, attaining the best SUN score among the evaluated baselines while sampling substantially faster than geometry-heavy alternatives. These results show that strong crystal modeling performance does not necessarily require full equivariance, provided that periodic geometry and chemical structure are incorporated in the right way. Overall, Crystalite offers a simple and efficient approach to crystal modeling and suggests that lightweight diffusion Transformers are a promising direction for scalable materials discovery. 
\bibliographystyle{plainnat}
\bibliography{refs}
\appendix

\clearpage
\section{Introduction to Materials}
\label{apdx:introduction_to_materials}
\subsection{Unit-cell representation of crystals}

\begin{figure}[H]
  \centering
  \includegraphics[width=0.9\linewidth]{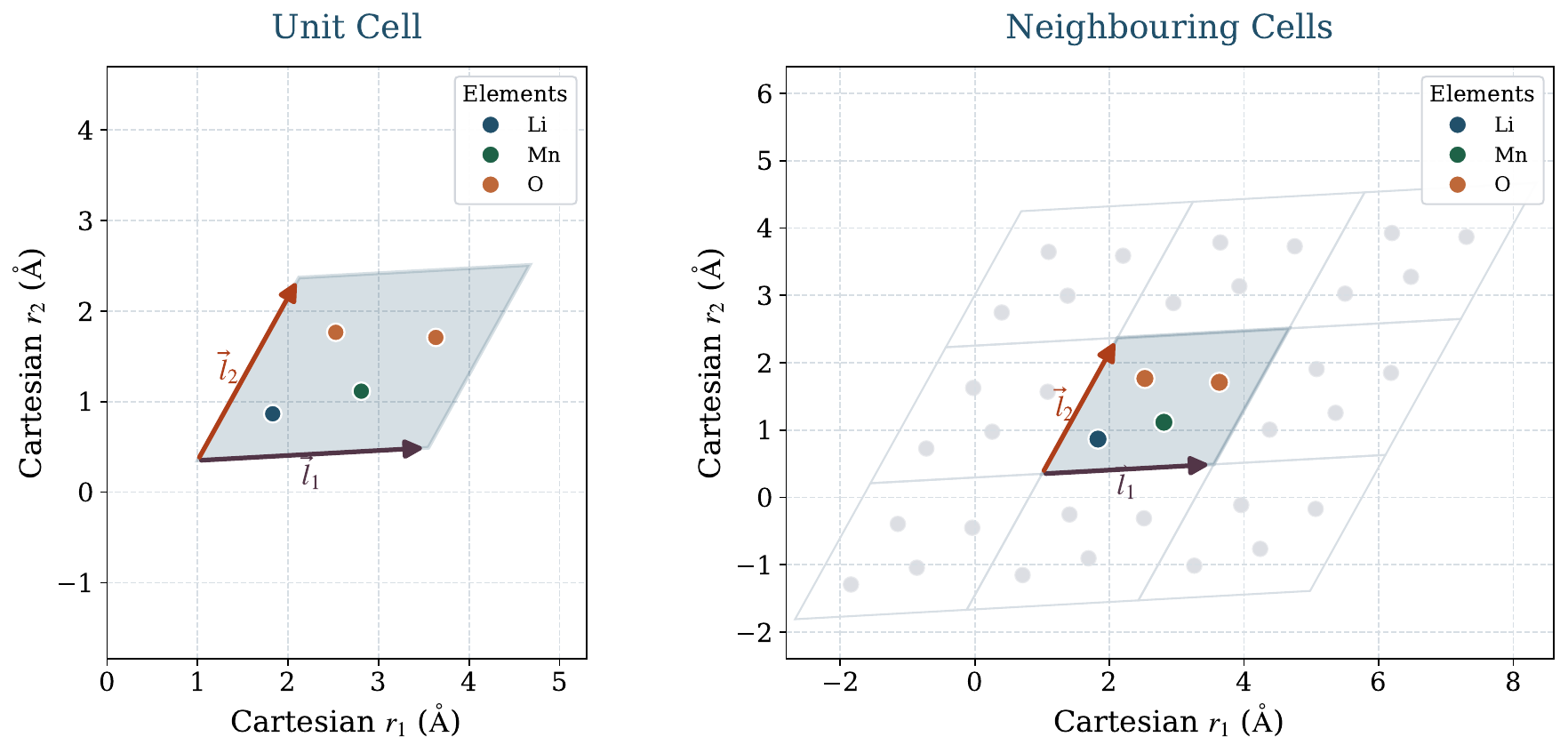}
  \caption{A crystal can be represented by a unit cell together with its periodic repetition under lattice translations.}
  \label{fig:unit_cell_supercell}
\end{figure}

A crystalline material is, ideally, an infinite periodic arrangement of atoms in three-dimensional space, as shown in Figure \ref{fig:unit_cell_supercell}. Rather than describing the full solid atom by atom, it suffices to specify a single \emph{unit cell} together with the rule that this cell repeats under integer translations of the lattice. This is the standard representation used throughout the paper.

Concretely, we represent a crystal with \(N\) atoms by the triple
\begin{equation}
\mathcal C = (\mathbf A,\mathbf F,\mathbf L),
\end{equation}
where \(\mathbf A \in \{0,1\}^{N\times N_Z}\) is the atom-type matrix, \(\mathbf F = [\mathbf f_1^\top;\dots;\mathbf f_N^\top] \in [0,1)^{N\times 3}\) contains the fractional coordinates, and \(\mathbf L \in \mathbb R^{3\times 3}\) is the lattice matrix. Each row of \(\mathbf A\) satisfies \(\mathbf A_i = \mathrm{onehot}(a_i)\) for some atomic species \(a_i \in \{1,\dots,N_Z\}\). The pair \((\mathbf A,\mathbf F)\) specifies the basis atoms inside the cell, while \(\mathbf L\) determines the geometry of the cell itself.

\begin{figure}[H]
  \centering
  \includegraphics[width=0.9\linewidth]{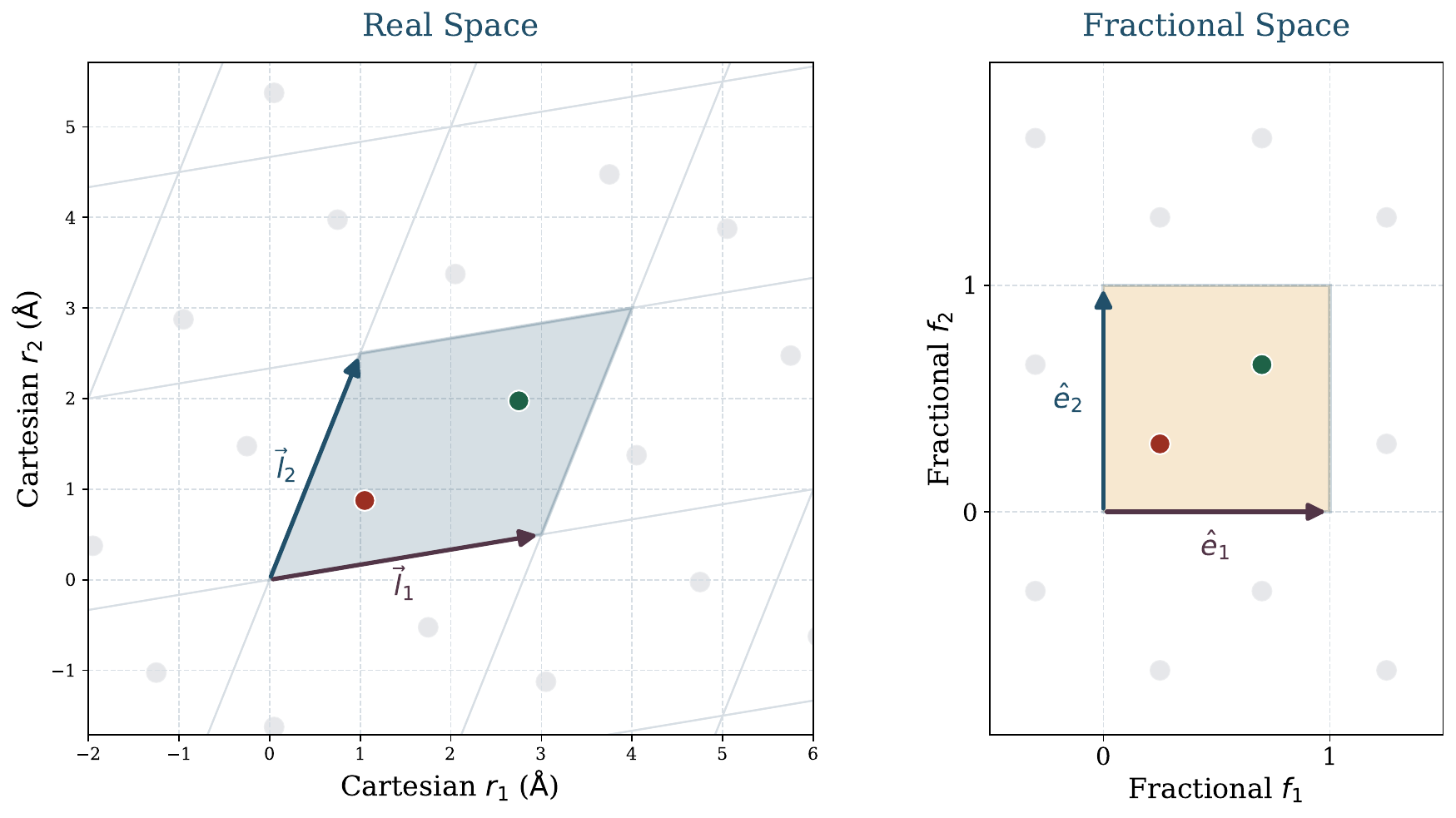}
  \caption{Fractional coordinates are defined in the unit cube and mapped to Cartesian space by the lattice matrix. Integer shifts in fractional coordinates correspond to lattice translations in real space.}
  \label{fig:fractional_coordinates}
\end{figure}

\paragraph{Fractional and Cartesian coordinates.}
We use fractional coordinates because they make periodicity explicit. Each row \(\mathbf f_i \in [0,1)^3\) gives the position of atom \(i\) relative to the lattice basis. Under the row-vector convention used in this paper, Cartesian coordinates are obtained by
\begin{equation}
\mathbf X = \mathbf F \mathbf L \in \mathbb R^{N\times 3},
\end{equation}
so that the Cartesian coordinate of atom \(i\) is the \(i\)-th row
\begin{equation}
\mathbf x_i = \mathbf f_i \mathbf L.
\end{equation}
Thus, \(\mathbf L\) controls the size and shape of the cell, while \(\mathbf F\) determines where atoms are placed inside it. Figure \ref{fig:fractional_coordinates} visualizes this transformation.

\paragraph{Periodic boundary conditions.}
Fractional coordinates live on the flat torus
\begin{equation}
\mathbb T^3 \cong (\mathbb R / \mathbb Z)^3,
\end{equation}
meaning that \(\mathbf f\) and \(\mathbf f + \mathbf n\) represent the same physical position for any \(\mathbf n \in \mathbb Z^3\). This is precisely the periodic boundary condition: atoms leaving one face of the unit cell re-enter through the opposite face.

The full infinite crystal is therefore generated by translating each basis atom by all integer lattice shifts:
\begin{equation}
\mathbf x_{i,\mathbf n} = (\mathbf f_i + \mathbf n)\mathbf L,
\qquad
\mathbf n \in \mathbb Z^3.
\end{equation}
A finite unit-cell description thus implicitly defines the entire periodic material.

\paragraph{Wrapped residuals and metric-aware minimum-image geometry.}
Because fractional coordinates are periodic, geometric quantities must respect the torus structure. In the coordinate loss, we use the componentwise wrapped residual in fractional space,
\begin{equation}
\bm{\delta}^{\mathrm{wrap}}_{ij}
=
\operatorname{wrap}(\mathbf f_i - \mathbf f_j),
\qquad
\operatorname{wrap}(\mathbf u)=\mathbf u-\operatorname{round}(\mathbf u),
\end{equation}
so that each component of \(\bm{\delta}^{\mathrm{wrap}}_{ij}\) lies in \([-\tfrac{1}{2},\tfrac{1}{2})\). The associated Cartesian displacement and distance are
\begin{equation}
\mathbf r^{\mathrm{wrap}}_{ij} = \bm{\delta}^{\mathrm{wrap}}_{ij}\mathbf L,
\qquad
d^{\mathrm{wrap}}_{ij} = \|\mathbf r^{\mathrm{wrap}}_{ij}\|_2.
\end{equation}

In the Geometry Enhancement Module (GEM), however, we do not use componentwise wrapping. Instead, we use a metric-aware periodic-image search under the lattice metric. Writing
\begin{equation}
\mathbf G = \mathbf L \mathbf L^\top,
\end{equation}
and restricting the search to a finite set of lattice offsets \(\Omega_R = \{-R,\dots,R\}^3\), we define
\begin{equation}
\Delta \mathbf f^\star_{ij}
=
\arg\min_{\mathbf r \in \Omega_R}
(\mathbf f_i - \mathbf f_j + \mathbf r)\,
\mathbf G\,
(\mathbf f_i - \mathbf f_j + \mathbf r)^\top,
\end{equation}
with corresponding Cartesian displacement and distance
\begin{equation}
\mathbf r^\star_{ij} = \Delta \mathbf f^\star_{ij}\mathbf L,
\qquad
d^\star_{ij} = \|\mathbf r^\star_{ij}\|_2.
\end{equation}
For orthogonal cells these two constructions coincide, but for general non-orthogonal cells they need not be equivalent. Throughout the paper, we therefore distinguish between the wrapped fractional residual used in the coordinate loss and the metric-aware minimum-image geometry used in GEM. When we refer to minimum-image geometry, we mean the latter construction.

\subsection{Symmetries and representation non-uniqueness}

The same physical crystal can admit multiple equivalent representations. As a result, the target distribution over crystals should respect several symmetries. In the notation of the main paper, these can be expressed directly in terms of \((\mathbf A,\mathbf F,\mathbf L)\).

\paragraph{Permutation of atom indices.}
The ordering of atoms inside the unit cell is arbitrary. For any permutation matrix \(P \in \mathcal P_N\),
\begin{equation}
p(\mathbf A,\mathbf F,\mathbf L)
=
p(P\mathbf A,\; P\mathbf F,\; \mathbf L).
\label{eq:sym_atom_perm}
\end{equation}

\paragraph{Global rotation in Cartesian space.}
A rigid rotation of the entire crystal changes only the Cartesian frame, not the underlying material. Under our row-vector convention, this corresponds to right multiplication of the lattice matrix. For any rotation \(R \in SO(3)\),
\begin{equation}
p(\mathbf A,\mathbf F,\mathbf L)
=
p(\mathbf A,\; \mathbf F,\; \mathbf L R).
\label{eq:sym_lattice_rot}
\end{equation}

\paragraph{Permutation of the lattice basis.}
The choice of lattice basis vectors is not unique. Permuting the lattice basis while applying the inverse permutation to the fractional coordinates leaves the Cartesian crystal unchanged. For any \(S \in \mathcal P_3\),
\begin{equation}
p(\mathbf A,\mathbf F,\mathbf L)
=
p(\mathbf A,\; \mathbf F S^\top,\; S\mathbf L).
\label{eq:sym_basis_perm}
\end{equation}

\paragraph{Global translation on the torus.}
Shifting all fractional coordinates by the same torus element does not change the crystal. For any \(\mathbf t \in \mathbb T^3\),
\begin{equation}
p(\mathbf A,\mathbf F,\mathbf L)
=
p\!\big(\mathbf A,\; \operatorname{wrap}(\mathbf F + \mathbf 1 \mathbf t^\top),\; \mathbf L\big),
\label{eq:sym_frac_translation}
\end{equation}
where \(\mathbf 1 \in \mathbb R^N\) denotes the all-ones vector.

These symmetries motivate several of the design choices in Crystalite. In particular, we represent positions in fractional coordinates, use wrapped periodic residuals for coordinate denoising, use metric-aware minimum-image geometry in GEM, and apply random global translations during training to encourage approximate translation equivariance.

\clearpage
\section{Subatomic Tokenization of Atoms}
\label{apdx:subatomic_tokenization}
\subsection{Subatomic Tokenization}
\label{apdx:subatomic_tokenization_details}

\begin{figure}[t]
    \centering
    \includegraphics[width=0.7\linewidth]{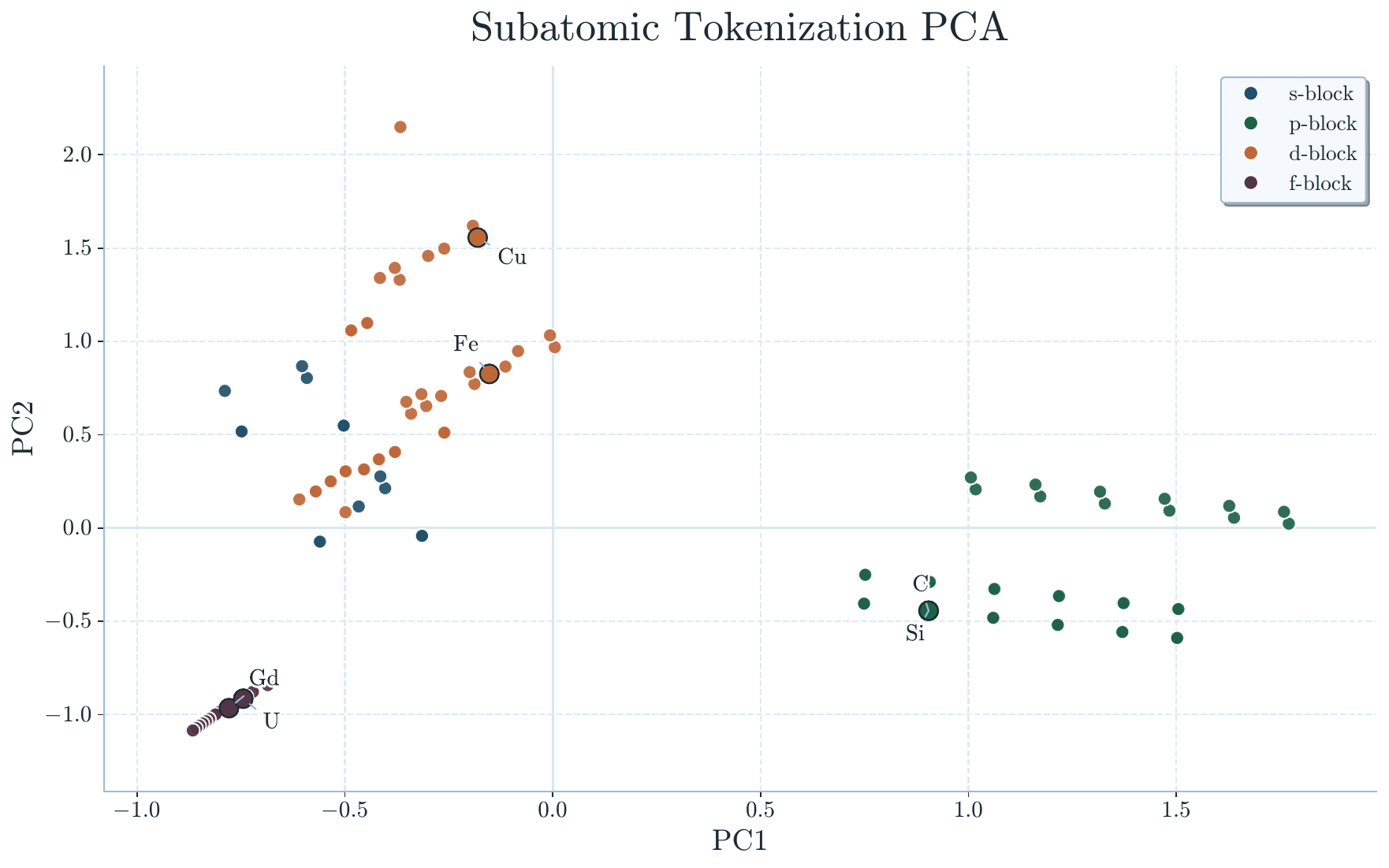}
    \caption{\textbf{Two-dimensional PCA projection of the subatomic tokens.} Each point corresponds to one supported element after projecting the balanced descriptors onto the first two principal components. The projection shows that the tokenization preserves meaningful chemical organization even after strong dimensionality reduction.}
    \label{fig:chem_pca_2d}
\end{figure}
We replace the usual one-hot atom identity with a continuous token that encodes basic chemical structure while still allowing deterministic decoding back to a valid element. The construction starts from simple periodic-table information and valence-shell occupancies, then standardizes and balances these features before optionally compressing them with PCA.

Let \(a_i \in \{1,\dots,N_Z\}\) denote the atomic number at site \(i\). For each supported element \(z \in \{1,\dots,N_Z\}\), we build a descriptor from four ingredients: its period, its group, its block, and its ground-state valence-shell occupancies. Concretely, let \(r(z)\in\{1,\dots,7\}\) be the period, \(g(z)\in\{0,\dots,18\}\) the group, where \(g(z)=0\) is reserved for \(f\)-block elements, and \(b(z)\in\{s,p,d,f\}\) the block. Let \((s_z,p_z,d_z,f_z)\) denote the corresponding valence occupancies from a fixed lookup table. We then define the raw descriptor
\begin{equation}
\mathbf d_z
=
\Big[
\mathrm{onehot}_{7}\!\big(r(z)-1\big),\;
\mathrm{onehot}_{19}\!\big(g(z)\big),\;
\mathrm{onehot}_{4}\!\big(b(z)\big),\;
s_z/2,\; p_z/6,\; d_z/10,\; f_z/14
\Big].
\label{eq:chem_token_raw}
\end{equation}
In our implementation this gives a \(34\)-dimensional vector, since
\[
7 + 19 + 4 + 4 = 34.
\]

Because these feature groups have different dimensionalities, we standardize each coordinate across the supported elements and then rebalance the groups so that large one-hot blocks do not dominate purely because they contain more entries. Let
\[
D=
\begin{bmatrix}
\mathbf d_1^\top\\
\vdots\\
\mathbf d_{N_Z}^\top
\end{bmatrix}
\in\mathbb R^{N_Z \times 34}
\]
collect the raw descriptors for all elements. We compute the featurewise mean and standard deviation,
\[
\bm\mu = \frac{1}{N_Z}\sum_{z=1}^{N_Z}\mathbf d_z,
\qquad
\bm\sigma = \mathrm{std}(D),
\]
and form the standardized descriptor
\begin{equation}
\tilde{\mathbf d}_z = (\mathbf d_z - \bm\mu)\oslash \bm\sigma,
\end{equation}
where \(\oslash\) denotes elementwise division. Any near-zero entry of \(\bm\sigma\) is replaced by \(1\) for numerical stability.

We next split \(\tilde{\mathbf d}_z\) into the four groups
\[
\text{period }(7), \qquad
\text{group }(19), \qquad
\text{block }(4), \qquad
\text{valence }(4),
\]
and rescale each group by the inverse square root of its dimensionality. If \(\tilde{\mathbf d}_z^{(G)}\) denotes the subvector corresponding to group \(G\), we define
\begin{equation}
\bar{\mathbf d}_z^{(G)} = |G|^{-1/2}\,\tilde{\mathbf d}_z^{(G)}.
\end{equation}
Concatenating the reweighted groups gives the balanced descriptor \(\bar{\mathbf d}_z\). The final raw token is then obtained by \(\ell_2\)-normalization,
\begin{equation}
\mathbf h_z
=
\frac{\bar{\mathbf d}_z}{\|\bar{\mathbf d}_z\|_2}.
\label{eq:chem_token_final}
\end{equation}

\begin{figure}[t]
    \centering
    \includegraphics[width=0.7\linewidth]{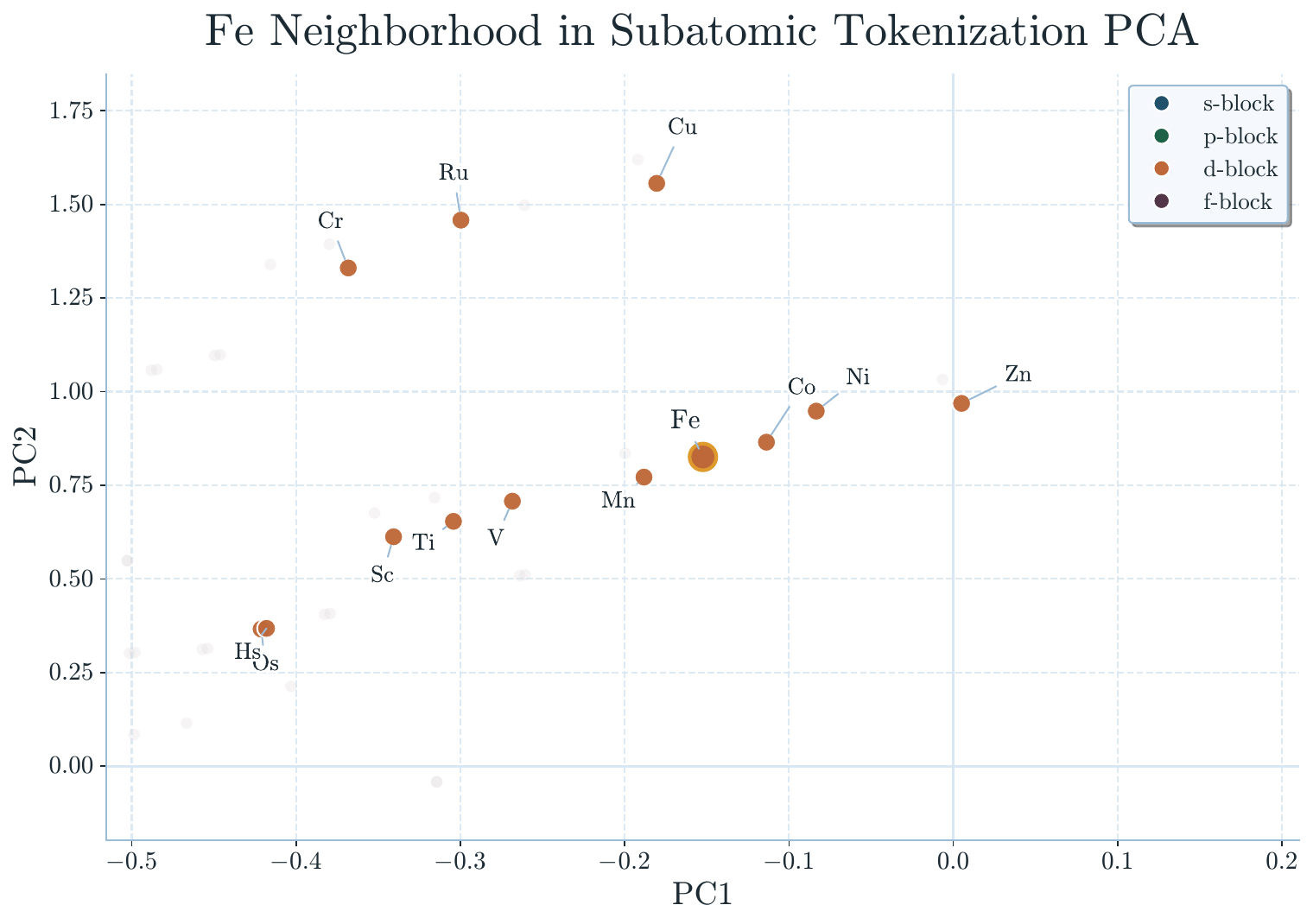}
    \caption{\textbf{Local neighborhood of Fe in the two-dimensional PCA space.} The plot highlights Fe together with its nearest elements in the projected representation, illustrating how the learned token geometry places chemically related species close to one another.}
    \label{fig:chem_pca_fe_neighbors}
\end{figure}

For a crystal with atomic numbers \((a_1,\dots,a_N)\), the atom-type channel becomes
\begin{equation}
\mathbf H
=
\begin{bmatrix}
\mathbf h_{a_1}^\top\\
\vdots\\
\mathbf h_{a_N}^\top
\end{bmatrix}
\in \mathbb R^{N \times d_H},
\end{equation}
with \(d_H=34\) in the raw representation.

When a lower-dimensional token is preferred, we apply PCA to the balanced descriptors. Let
\[
\bar{D}
=
\begin{bmatrix}
\bar{\mathbf d}_1^\top\\
\vdots\\
\bar{\mathbf d}_{N_Z}^\top
\end{bmatrix}
\in \mathbb R^{N_Z \times 34},
\]
and let \(U_d \in \mathbb R^{34 \times d}\) contain the top \(d\) principal directions. Each element is then represented by
\begin{equation}
\mathbf p_z = \bar{\mathbf d}_z U_d \in \mathbb R^d,
\qquad
\mathbf h_z^{\mathrm{PCA}} = \frac{\mathbf p_z}{\|\mathbf p_z\|_2}.
\end{equation}
This gives a compressed tokenization with \(d_H=d\). A two-dimensional PCA projection of the element tokens is shown in Figure~\ref{fig:chem_pca_2d}. Even in two dimensions, the representation retains visible chemical structure. Figure~\ref{fig:chem_pca_fe_neighbors} shows the local neighborhood of Fe in this projected space, which provides an intuitive view of how chemically related elements cluster around it.

Finally, both the raw and PCA-compressed tokens can be decoded deterministically by nearest-prototype matching. Given a predicted continuous token \(\hat{\mathbf h}_i\), we assign the atomic species as
\begin{equation}
\hat a_i
=
\arg\max_{z\in\{1,\dots,N_Z\}}
\langle \hat{\mathbf h}_i,\mathbf h_z^\star \rangle,
\end{equation}
where \(\mathbf h_z^\star\) is either the raw prototype \(\mathbf h_z\) or the PCA-compressed prototype \(\mathbf h_z^{\mathrm{PCA}}\). Since all prototypes are normalized, this is equivalent to cosine-similarity decoding.

\clearpage
\section{Crystalite Architecture}
\label{apdx:architecture}
\begin{figure}[H]
    \centering
    \includegraphics[width=0.95\linewidth]{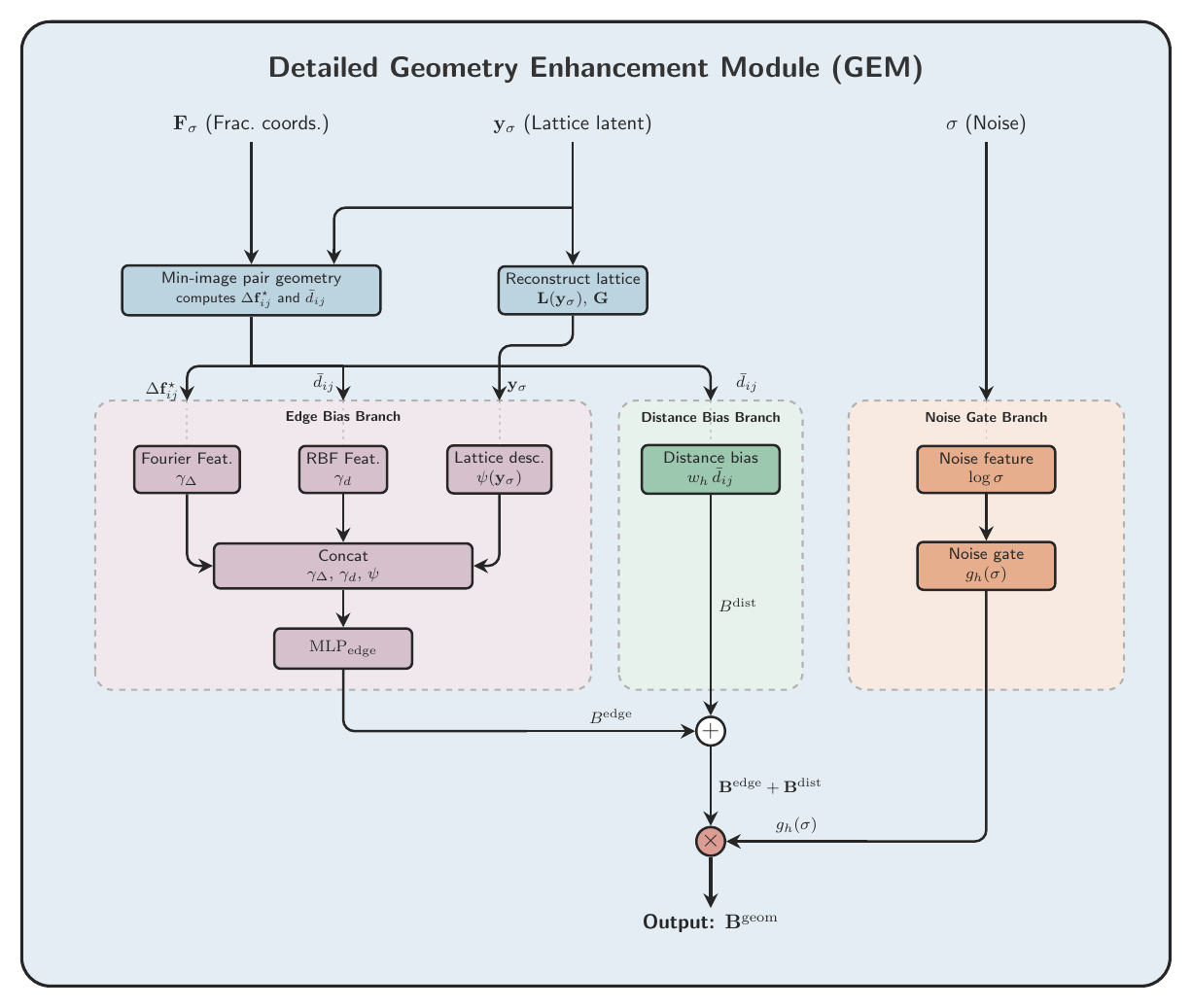}
    \caption{
        Detailed view of the Geometry Enhancement Module (GEM). Starting from the current fractional coordinates and lattice parameters, GEM computes periodic pairwise geometry under minimum-image conventions, converts it into distance and edge-aware attention biases, and injects the resulting signal additively into the attention logits.
    }
    \label{fig:gem_module_detailed}
\end{figure}

This appendix provides a more detailed description of Crystalite using the notation of the main text. Recall that a crystal is represented as
\[
\mathcal C = (\mathbf A,\mathbf F,\mathbf L),
\]
and that the diffusion model operates on the continuous state
\[
(\mathbf H,\mathbf F,\mathbf y),
\]
where \(\mathbf H\) denotes atom-type features obtained from \(\mathbf A\), and \(\mathbf y\in\mathbb R^6\) is the lower-triangular lattice parameterization satisfying \(\mathbf L=\mathbf L(\mathbf y)\). In DNG, \(\mathbf H\) is represented with PCA-compressed Subatomic Tokenization; in CSP, atom-type features are fixed atomic-number features. Figure~\ref{fig:crystalite_architecture} gives an overview of the full architecture, while Figure~\ref{fig:gem_module_detailed} illustrates the Geometry Enhancement Module (GEM).

\subsection{Tokenization and input embeddings}

Each atomic site \(i\) contributes one token to the Transformer sequence. The atom-type feature vector \(\mathbf H_i \in \mathbb R^{d_H}\) is first mapped to the model dimension through a learned embedder \(E_H\),
\begin{equation}
\mathbf h_i^{H} = E_H(\mathbf H_i),
\end{equation}
where \(E_H\) is implemented as a two-layer MLP with SiLU activation acting directly on the continuous atom token:
\[
E_H:\ \mathbb R^{d_H} \to \mathbb R^d,
\qquad
\mathrm{Linear}(d_H,d)\; \rightarrow\; \mathrm{SiLU}\; \rightarrow\; \mathrm{Linear}(d,d).
\]
The corresponding fractional coordinate \(\mathbf f_i \in [0,1)^3\) is embedded separately through
\begin{equation}
\mathbf h_i^{F} = E_F(\mathbf f_i) = \operatorname{MLP}_F\!\big(\gamma_F(\mathbf f_i)\big),
\end{equation}
where \(\gamma_F\) denotes a deterministic Fourier feature map. Concretely, we use sinusoidal features at multiple frequencies,
\[
\gamma_F(\mathbf f_i)
=
\big[
\sin(2\pi \ell \mathbf f_i),\,
\cos(2\pi \ell \mathbf f_i)
\big]_{\ell=1}^{n_F},
\]
followed by a two-layer MLP with SiLU activation. Thus \(E_F\) has the form
\[
E_F:\ \mathbb R^{6n_F} \to \mathbb R^d,
\qquad
\mathrm{Linear}(6n_F,d)\; \rightarrow\; \mathrm{SiLU}\; \rightarrow\; \mathrm{Linear}(d,d),
\]
with \(n_F=32\) in the base configuration. The resulting atom token is
\begin{equation}
\mathbf t_i^{\mathrm{atom}} = E_H(\mathbf H_i) + E_F(\mathbf f_i).
\end{equation}

The lattice is represented by a single global token. The lattice latent \(\mathbf y \in \mathbb R^6\) is the lower-triangular parameterization introduced in Eq.~\eqref{eq:lattice_param}, and is embedded through
\begin{equation}
\mathbf t^{\mathrm{lat}} = E_{\mathrm{lat}}(\mathbf y),
\end{equation}
where \(E_{\mathrm{lat}}\) is implemented as a two-layer MLP with SiLU activation acting directly on \(\mathbf y\):
\[
E_{\mathrm{lat}}:\ \mathbb R^6 \to \mathbb R^d,
\qquad
\mathrm{Linear}(6,d)\; \rightarrow\; \mathrm{SiLU}\; \rightarrow\; \mathrm{Linear}(d,d).
\]
For a crystal with \(N\) atoms, the initial Transformer sequence is therefore
\begin{equation}
\mathbf T^{(0)}
=
\big[
\mathbf t_1^{\mathrm{atom}}, \dots, \mathbf t_N^{\mathrm{atom}}, \mathbf t^{\mathrm{lat}}
\big]
\in \mathbb R^{(N+1)\times d}.
\end{equation}
Thus Crystalite uses one token per atom, together with one additional token that summarizes the global unit-cell geometry.

The diffusion noise level is embedded through the standard EDM noise coordinate
\begin{equation}
c_{\mathrm{noise}}(\sigma) = \tfrac{1}{4}\log \sigma,
\end{equation}
followed by a learned embedder \(E_\sigma\), giving a conditioning vector
\begin{equation}
\mathbf c_\sigma = E_\sigma\!\big(c_{\mathrm{noise}}(\sigma)\big).
\end{equation}
This conditioning is injected into every Transformer block through adaptive layer normalization (AdaLN).

The token sequence is then processed by a standard Transformer trunk with stacked self-attention and feed-forward blocks. Writing \(\mathbf T^{(k)}\) for the token sequence entering block \(k\), the update can be written schematically as
\begin{align}
\mathbf T^{(k+\frac{1}{2})}
&=
\mathbf T^{(k)}
+
\operatorname{MHA}^{(k)}\!\big(\mathbf T^{(k)};\mathbf c_\sigma,\widetilde{\mathbf B}^{(k)}\big),\\
\mathbf T^{(k+1)}
&=
\mathbf T^{(k+\frac{1}{2})}
+
\operatorname{MLP}^{(k)}\!\big(\mathbf T^{(k+\frac{1}{2})};\mathbf c_\sigma\big),
\end{align}
where \(\widetilde{\mathbf B}^{(k)}\) denotes the optional additive attention bias produced by GEM. When GEM is disabled, \(\widetilde{\mathbf B}^{(k)}=0\) and the model reduces to a standard AdaLN-conditioned diffusion Transformer.

After the final block, shallow output heads map the updated atom tokens to denoised atom-type and coordinate predictions, and the lattice token to the denoised lattice latent:
\begin{equation}
\hat{\mathbf H}_i = D_H(\mathbf t_i^{(K)}),
\qquad
\hat{\mathbf f}_i = D_F(\mathbf t_i^{(K)}),
\qquad
\hat{\mathbf y} = D_{\mathrm{lat}}(\mathbf t^{(K)}_{\mathrm{lat}}).
\end{equation}
Thus atom-wise quantities are predicted from the site tokens, while the global lattice parameters are predicted from the lattice token.
\subsection{Geometry Enhancement Module (GEM)}
\label{apdx:GEM}

GEM augments self-attention with pairwise geometric biases derived from the current crystal geometry. It does not change the tokenization or prediction heads; instead, it modifies the attention logits through an additive bias tensor.

Given the current fractional coordinates \(\mathbf F\) and lattice latent \(\mathbf y\), GEM first reconstructs the lattice matrix \(\mathbf L(\mathbf y)\) and computes pairwise minimum-image geometry under periodic boundary conditions. Let
\begin{equation}
\mathbf G(\mathbf y) = \mathbf L(\mathbf y)\mathbf L(\mathbf y)^\top
\end{equation}
denote the corresponding metric tensor. For each pair of atoms \((i,j)\), we consider periodic offsets \(\mathbf r \in \Omega_R = \{-R,\dots,R\}^3\) and define
\begin{equation}
\Delta \mathbf f_{ij}(\mathbf r) = \mathbf f_i - \mathbf f_j + \mathbf r.
\end{equation}
The minimum-image displacement is then chosen as
\begin{equation}
\Delta \mathbf f_{ij}^{\star}
=
\arg\min_{\mathbf r \in \Omega_R}
\Delta \mathbf f_{ij}(\mathbf r)\,
\mathbf G(\mathbf y)\,
\Delta \mathbf f_{ij}(\mathbf r)^\top,
\end{equation}
with corresponding Cartesian distance
\begin{equation}
d_{ij}
=
\big\|
\Delta \mathbf f_{ij}^{\star}\mathbf L(\mathbf y)
\big\|_2.
\end{equation}
In practice, this distance is normalized by a characteristic cell scale \(s(\mathbf y)\), yielding \(\bar d_{ij}=d_{ij}/s(\mathbf y)\).

From this pairwise geometry, GEM builds two additive bias terms. The first is a distance bias,
\begin{equation}
B^{\mathrm{dist}}_{hij} = \alpha_h\,\bar d_{ij},
\qquad
\alpha_h \le 0,
\end{equation}
which acts as a learnable locality prior for each attention head \(h\). The second is an edge-aware bias produced by a small MLP acting on periodic pairwise features,
\begin{equation}
\phi_{ij}
=
\big[
\gamma_{\Delta}(\Delta \mathbf f_{ij}^{\star}),
\gamma_d(\bar d_{ij}),
\psi(\mathbf y)
\big],
\qquad
B^{\mathrm{edge}}_{hij}
=
\operatorname{MLP}_{\mathrm{edge}}(\phi_{ij})_h,
\end{equation}
where \(\gamma_{\Delta}\) and \(\gamma_d\) denote Fourier/RBF feature maps and \(\psi(\mathbf y)\) is a low-dimensional lattice descriptor.

The two branches are combined, optionally modulated by a noise-dependent gate,
\begin{equation}
B_{hij}^{(k)}
=
g_h(\sigma)\Big(
B^{\mathrm{dist}}_{hij}
+
B^{\mathrm{edge}}_{hij}
\Big),
\end{equation}
and then expanded from atom pairs to the full token sequence by leaving lattice-token interactions unbiased:
\begin{equation}
\widetilde{\mathbf B}^{(k)}_h
=
\begin{bmatrix}
\mathbf B_h^{(k)} & \mathbf 0\\
\mathbf 0^\top & 0
\end{bmatrix}.
\end{equation}
Finally, this bias is added directly to the attention logits,
\begin{equation}
\operatorname{Attn}(Q,K,V)
=
\operatorname{softmax}\!\left(
\frac{QK^\top}{\sqrt d}
+
\widetilde{\mathbf B}^{(k)}
\right)V.
\end{equation}

This construction lets Crystalite inject periodic geometric information directly into attention while preserving the simplicity of a standard Transformer backbone. When GEM is disabled, the model uses the same tokenization, diffusion objective, and output heads, but with \(\widetilde{\mathbf B}^{(k)}=0\).
\newpage
\subsection{Task-specific model configurations}

The same Crystalite model family is used for both \textit{de novo} generation and crystal structure prediction, but the reported experiments use task-specific configurations. Table~\ref{tab:crystalite_base_config} summarizes the configurations used for the main DNG and CSP results. The Alex-MP-20 CSP experiment uses the same configuration as MP-20.

\begin{table}[H]
    \centering
    \footnotesize
    \setlength{\tabcolsep}{3pt}
    \renewcommand{\arraystretch}{1.08}
    \caption{Task-specific Crystalite configurations used for the main DNG and CSP experiments. The Subatomic Tokenization row uses \(\checkmark\) when PCA-compressed Subatomic Tokenization is used and \(\times\) when atomic-number features are used instead. Alex-MP-20 CSP uses the same configuration as MP-20. For CSP, the effective atom-type loss weight is zero because the composition is fixed.}
    \label{tab:crystalite_base_config}

    \begin{subtable}[t]{\textwidth}
        \centering
        \caption{Architecture}
        \begin{tabular}{@{}p{0.31\linewidth}p{0.20\linewidth}p{0.20\linewidth}p{0.20\linewidth}@{}}
            \toprule
            \textbf{Component} & \textbf{DNG MP-20} & \shortstack[l]{\textbf{CSP MP-20}\\\textbf{Alex-MP-20}} & \textbf{CSP MPTS-52} \\
            \midrule
            Generated / denoised channels & types, coords, lattice & coords, lattice & coords, lattice \\
            Subatomic Tokenization & \(\checkmark\) & \(\times\) & \(\times\) \\
            Atom-type feature dimension & 16 & 95 & 95 \\
            Transformer width \(d\) & 512 & 1024 & 1024 \\
            Transformer layers & 14 & 14 & 14 \\
            Attention heads & 16 & 16 & 16 \\
            Dropout / attn. dropout & 0 / 0 & 0 / 0 & 0 / 0 \\
            Coordinate embedding & Fourier, 32 freqs. & Fourier, 32 freqs. & Fourier, 32 freqs. \\
            Coordinate head & direct fractional head & direct fractional head & direct fractional head \\
            Lattice representation & lower-triangular & lower-triangular & lower-triangular \\
            Lattice embedder & MLP & MLP & MLP \\
            GEM distance bias & disabled & enabled & disabled \\
            GEM edge-aware bias & enabled & enabled & enabled \\
            Edge-bias freqs. / hidden / RBF & 12 / 256 / 32 & 12 / 256 / 32 & 12 / 256 / 32 \\
            GEM sharing & shared across layers & per-layer & per-layer \\
            PBC search radius \(R\) & 1 & 1 & 1 \\
            Max atoms per cell & 20 & 20 & 52 \\
            \bottomrule
        \end{tabular}
    \end{subtable}

    \vspace{0.75em}

    \begin{subtable}[t]{\textwidth}
        \centering
        \caption{Training and sampling}
        \begin{tabular}{@{}p{0.31\linewidth}p{0.20\linewidth}p{0.20\linewidth}p{0.20\linewidth}@{}}
            \toprule
            \textbf{Component} & \textbf{DNG MP-20} & \shortstack[l]{\textbf{CSP MP-20}\\\textbf{Alex-MP-20}} & \textbf{CSP MPTS-52} \\
            \midrule
            Batch size & 128 & 128 & 128 \\
            Learning rate & \(10^{-4}\) & \(10^{-4}\) & \(10^{-4}\) \\
            Weight decay & 0 & 0 & 0 \\
            LR warmup & 1000 steps & 1000 steps & 1000 steps \\
            Training steps & \(2.5\times 10^6\) & \(5.0\times 10^6\) & \(3.0\times 10^6\) \\
            EMA decay & 0.9999 & 0.99999 & 0.9999 \\
            Precision & bfloat16 & bfloat16 & bfloat16 \\
            EDM \((P_{\mathrm{mean}},P_{\mathrm{std}})\) & \((-1.2,\,1.2)\) & \((-1.2,\,1.2)\) & \((-1.2,\,1.2)\) \\
            \(\sigma_{\mathrm{data}}\) \((H,F,\mathrm{lat})\) & \((0.3,\,0.3,\,0.3)\) & \((1.0,\,0.3,\,0.3)\) & \((1.0,\,0.3,\,0.3)\) \\
            Effective loss weights \((\lambda_H,\lambda_F,\lambda_{\mathrm{lat}})\) & \((1,\,50,\,5)\) & \((0,\,20,\,10)\) & \((0,\,20,\,10)\) \\
            Coordinate loss & wrapped fractional MSE & wrapped fractional MSE & wrapped fractional MSE \\
            Sampling steps & 150 & 400 & 400 \\
            \((\rho,S_{\mathrm{churn}},S_{\mathrm{noise}})\) & \((7,\,60,\,1.003)\) & \((7,\,30,\,1.003)\) & \((7,\,30,\,1.003)\) \\
            \((S_{\min},S_{\max})\) & \((0,\,999)\) & \((0,\,999)\) & \((0,\,999)\) \\
            Anti-annealing \((H,F,\mathrm{lat})\) & \((0,\,10,\,10)\) & \((0,\,4,\,4)\) & \((0,\,4,\,4)\) \\
            Sampling weights & EMA & EMA & EMA \\
            \bottomrule
        \end{tabular}
    \end{subtable}
\end{table}

The broader implementation supports additional tokenizations, embedding variants, and GEM configurations, but Table~\ref{tab:crystalite_base_config} gives the task-specific settings used for the main reported results.

\clearpage
\section{EDM Training Details}
\label{apdx:edm}
\paragraph{EDM noising and preconditioning.}
At each training step, we sample a noise level according to
\begin{equation}
\log \sigma \sim \mathcal{N}(P_{\mathrm{mean}},P_{\mathrm{std}}^2).
\end{equation}
Following the notation of the main text, the diffusion model operates on the continuous crystal state
\[
(\mathbf H,\mathbf F,\mathbf y),
\]
where \(\mathbf H\) denotes atom-type features, \(\mathbf F \in [0,1)^{N\times 3}\) the fractional coordinates, and \(\mathbf y \in \mathbb R^6\) the lattice latent. In DNG, \(\mathbf H\) is represented with Subatomic Tokenization; in CSP, atom-type features are fixed by the known composition.

The atom-token and lattice channels are noised directly in Euclidean space, while the coordinate channel is noised in a centered representation. Concretely, we define
\begin{equation}
\mathbf F_{\mathrm c} = \mathbf F - \tfrac{1}{2},
\end{equation}
and then sample
\begin{equation}
\widetilde{\mathbf H} = \mathbf H + \sigma \bm{\varepsilon}_H,
\qquad
\widetilde{\mathbf F}_{\mathrm c} = \mathbf F_{\mathrm c} + \sigma \bm{\varepsilon}_F,
\qquad
\widetilde{\mathbf y} = \mathbf y + \sigma \bm{\varepsilon}_{\mathrm{lat}},
\end{equation}
with independent Gaussian noise terms. Before the coordinate embedder, the noisy centered coordinates are shifted back and wrapped into the unit cube,
\begin{equation}
\widetilde{\mathbf F}_{\mathrm{in}}
=
\operatorname{mod1}\!\left(\widetilde{\mathbf F}_{\mathrm c} + \tfrac{1}{2}\right),
\qquad
\operatorname{mod1}(\mathbf u)=\mathbf u-\lfloor \mathbf u \rfloor.
\end{equation}
The noise level is provided to the Transformer through the usual EDM conditioning scalar
\begin{equation}
c_{\mathrm{noise}}(\sigma)=\tfrac{1}{4}\log\sigma.
\end{equation}

For each channel \(u\in\{H,F,\mathrm{lat}\}\), we use the standard EDM preconditioning coefficients
\begin{equation}
c_{\mathrm{skip},u}(\sigma)=\frac{\sigma_{\mathrm{data},u}^{2}}{\sigma^{2}+\sigma_{\mathrm{data},u}^{2}},
\qquad
c_{\mathrm{out},u}(\sigma)=\frac{\sigma\,\sigma_{\mathrm{data},u}}{\sqrt{\sigma^{2}+\sigma_{\mathrm{data},u}^{2}}},
\qquad
c_{\mathrm{in},u}(\sigma)=\frac{1}{\sqrt{\sigma^{2}+\sigma_{\mathrm{data},u}^{2}}}.
\end{equation}
In our implementation, the atom-token and lattice channels are scaled by \(c_{\mathrm{in},u}(\sigma)\) before being passed to the network, whereas the coordinate channel is passed as wrapped fractional coordinates \(\widetilde{\mathbf F}_{\mathrm{in}}\). Denoting the raw network outputs by \(\mathbf R_H\), \(\mathbf R_F\), and \(\mathbf R_{\mathrm{lat}}\), the corresponding denoised predictions are
\begin{equation}
\hat{\mathbf H}
=
c_{\mathrm{skip},H}(\sigma)\,\widetilde{\mathbf H}
+
c_{\mathrm{out},H}(\sigma)\,\mathbf R_H,
\end{equation}
\begin{equation}
\hat{\mathbf F}_{\mathrm c}
=
c_{\mathrm{skip},F}(\sigma)\,\widetilde{\mathbf F}_{\mathrm c}
+
c_{\mathrm{out},F}(\sigma)\,\mathbf R_F,
\end{equation}
\begin{equation}
\hat{\mathbf y}
=
c_{\mathrm{skip},\mathrm{lat}}(\sigma)\,\widetilde{\mathbf y}
+
c_{\mathrm{out},\mathrm{lat}}(\sigma)\,\mathbf R_{\mathrm{lat}}.
\end{equation}
For the coordinate loss, we map the centered prediction back to fractional coordinates,
\begin{equation}
\hat{\mathbf F}
=
\operatorname{mod1}\!\left(\hat{\mathbf F}_{\mathrm c}+\tfrac{1}{2}\right),
\end{equation}
and then compute the wrapped fractional residual
\[
\Delta_i=\mathrm{wrap}(\hat{\mathbf f}_i-\mathbf f_i),
\qquad
\mathrm{wrap}(\mathbf u)=\mathbf u-\mathrm{round}(\mathbf u),
\]
so that each component lies in \([-\tfrac12,\tfrac12)\). This is a torus-aware residual in fractional space, not the metric-aware minimum-image displacement used in GEM.

Finally, the EDM loss weights are
\begin{equation}
w_u(\sigma)
=
\frac{\sigma^{2}+\sigma_{\mathrm{data},u}^{2}}{(\sigma\,\sigma_{\mathrm{data},u})^{2}},
\qquad
u\in\{H,F,\mathrm{lat}\}.
\end{equation}
These are the weights used in the channel-wise training objective described in the main text.

\subsection{Channel-wise anti-annealing during sampling}
\label{apdx:anti_annealing}

We write the sampler state at step \(i\) as
\[
\mathbf z_i = (\mathbf H_i,\mathbf F_i,\mathbf y_i),
\qquad i=0,\dots,N,
\]
along a decreasing EDM noise schedule
\[
\sigma_0 > \sigma_1 > \cdots > \sigma_{N-1} > \sigma_N = 0,
\]
with
\begin{equation}
\sigma_i
=
\left(
\sigma_{\max}^{1/\rho}
+
\frac{i}{N-1}
\left(
\sigma_{\min}^{1/\rho}
-
\sigma_{\max}^{1/\rho}
\right)
\right)^{\rho},
\qquad i=0,\dots,N-1.
\end{equation}

As in EDM, we optionally apply churn at step \(i\), defining
\begin{equation}
\bar{\sigma}_i = (1+\gamma_i)\sigma_i,
\end{equation}
and the corresponding perturbed state
\begin{equation}
(\bar{\mathbf H}_i,\bar{\mathbf F}_i,\bar{\mathbf y}_i)
=
(\mathbf H_i,\mathbf F_i,\mathbf y_i)
+
\sqrt{\bar{\sigma}_i^2-\sigma_i^2}\,
(\bm\varepsilon_i^{H},\bm\varepsilon_i^{F},\bm\varepsilon_i^{y}),
\end{equation}
where the noise tensors have the appropriate shapes.

We then evaluate the denoiser at \(\bar{\sigma}_i\),
\begin{equation}
(\mathbf H_i^{\mathrm{den}},\mathbf F_i^{\mathrm{den}},\mathbf y_i^{\mathrm{den}})
=
D_\theta(\bar{\mathbf H}_i,\bar{\mathbf F}_i,\bar{\mathbf y}_i,\bar{\sigma}_i).
\end{equation}
The corresponding EDM drifts are
\begin{align}
\mathbf d_i^{H}
&=
\frac{\bar{\mathbf H}_i-\mathbf H_i^{\mathrm{den}}}{\bar{\sigma}_i}, \\
\mathbf d_i^{F}
&=
\frac{\operatorname{wrap}(\bar{\mathbf F}_i-\mathbf F_i^{\mathrm{den}})}{\bar{\sigma}_i}, \\
\mathbf d_i^{y}
&=
\frac{\bar{\mathbf y}_i-\mathbf y_i^{\mathrm{den}}}{\bar{\sigma}_i},
\end{align}
where \(\operatorname{wrap}(\mathbf u)=\mathbf u-\operatorname{round}(\mathbf u)\) is applied elementwise to respect periodicity in fractional coordinates.

To anti-anneal a selected channel \(q\in\{H,F,y\}\), we introduce an auxiliary Karras schedule
\begin{equation}
\tilde{\sigma}_i^{(q)}
=
\left(
\sigma_{\max}^{1/\rho_q^{\mathrm{AA}}}
+
\frac{i}{N-1}
\left(
\sigma_{\min}^{1/\rho_q^{\mathrm{AA}}}
-
\sigma_{\max}^{1/\rho_q^{\mathrm{AA}}}
\right)
\right)^{\rho_q^{\mathrm{AA}}},
\qquad i=0,\dots,N-1.
\end{equation}
Writing
\[
\Delta_i = \sigma_i-\sigma_{i+1},
\qquad
\tilde{\Delta}_i^{(q)}=\tilde{\sigma}_i^{(q)}-\tilde{\sigma}_{i+1}^{(q)},
\]
we define the anti-annealing factor
\begin{equation}
\alpha_i^{(q)}
=
\max\!\left(
1,\;
\frac{\tilde{\Delta}_i^{(q)}}{\Delta_i}
\right).
\end{equation}
If anti-annealing is disabled for channel \(q\), we set \(\alpha_i^{(q)}=1\). For fractional coordinates, we may additionally cap this factor,
\begin{equation}
\alpha_i^{(F)} \leftarrow \min\!\bigl(\alpha_i^{(F)},\alpha_{\max}\bigr).
\end{equation}

Let
\[
\mathbf z^{(H)}=\mathbf H,
\qquad
\mathbf z^{(F)}=\mathbf F,
\qquad
\mathbf z^{(y)}=\mathbf y.
\]
The Euler predictor step is then
\begin{equation}
\mathbf z_{i+1}^{(q),\mathrm{pred}}
=
\bar{\mathbf z}_i^{(q)}
+
(\sigma_{i+1}-\bar{\sigma}_i)\,
\alpha_i^{(q)}\,
\mathbf d_i^{(q)},
\qquad q\in\{H,F,y\}.
\end{equation}

When \(\sigma_{i+1}>0\), we apply the usual Heun correction. We first evaluate the denoiser at the predicted state,
\begin{equation}
(\mathbf H_{i+1}^{\mathrm{den}},\mathbf F_{i+1}^{\mathrm{den}},\mathbf y_{i+1}^{\mathrm{den}})
=
D_\theta(
\mathbf H_{i+1}^{\mathrm{pred}},
\mathbf F_{i+1}^{\mathrm{pred}},
\mathbf y_{i+1}^{\mathrm{pred}},
\sigma_{i+1}),
\end{equation}
and define corrected drifts
\begin{align}
\mathbf d_{i+1}^{H}
&=
\frac{\mathbf H_{i+1}^{\mathrm{pred}}-\mathbf H_{i+1}^{\mathrm{den}}}{\sigma_{i+1}}, \\
\mathbf d_{i+1}^{F}
&=
\frac{\operatorname{wrap}(\mathbf F_{i+1}^{\mathrm{pred}}-\mathbf F_{i+1}^{\mathrm{den}})}{\sigma_{i+1}}, \\
\mathbf d_{i+1}^{y}
&=
\frac{\mathbf y_{i+1}^{\mathrm{pred}}-\mathbf y_{i+1}^{\mathrm{den}}}{\sigma_{i+1}}.
\end{align}
The final Heun update becomes
\begin{equation}
\mathbf z_{i+1}^{(q)}
=
\bar{\mathbf z}_i^{(q)}
+
(\sigma_{i+1}-\bar{\sigma}_i)\,
\alpha_i^{(q)}\,
\frac{\mathbf d_i^{(q)}+\mathbf d_{i+1}^{(q)}}{2},
\qquad q\in\{H,F,y\}.
\end{equation}
At the terminal step, where \(\sigma_{i+1}=0\), we simply use the predictor:
\begin{equation}
\mathbf z_{i+1}^{(q)} = \mathbf z_{i+1}^{(q),\mathrm{pred}}.
\end{equation}

In this form, anti-annealing is a channel-wise rescaling of the EDM drift. Equivalently, it introduces a channel-dependent time warp: channels with \(\alpha_i^{(q)}>1\) are driven more aggressively toward their denoised predictions, while the denoiser itself and the underlying EDM schedule remain unchanged.

\clearpage
\section{Evaluation Details}
\label{apdx:eval}
\subsection{De novo generation (DNG)}
\label{apdx:evaluation_metrics}

For \textit{de novo} generation, we sample
\[
N_{\mathrm{gen}} = 10{,}000
\]
crystals and decode them into periodic structures
\[
\mathcal{G}=\{\mathcal C_1,\dots,\mathcal C_{N_{\mathrm{gen}}}\}.
\]
We report four groups of metrics: validity, uniqueness and novelty, distribution matching, and thermodynamic competitiveness.

\paragraph{Validity.}
We report composition validity, structure validity, and overall validity separately.

Composition validity is evaluated with SMACT \citep{Davies2019}. For each generated crystal, the stoichiometry is reduced to its primitive integer ratio, after which oxidation-state assignments, charge neutrality, and the Pauling electronegativity criterion are checked. Unary systems and all-metal alloys are handled in the standard way used in prior crystal-generation work.

Structure validity is implemented as a small pipeline rather than as a single geometric test. Before constructing a pymatgen \texttt{Structure}, the evaluator applies a safe-wrapper prefilter that rejects malformed decoded samples, including invalid atomic numbers and implausible lattice angles. The code then attempts to construct a periodic structure and marks the sample as structurally invalid if this fails, if lattice parameters or coordinates are non-finite, if lattice lengths are negative, or if the resulting cell volume is smaller than \(0.1\,\text{\AA}^3\). Only samples that survive these checks reach the final geometric validity test, which requires both
\begin{equation}
\mathrm{vol}(\mathcal C)\ge 0.1\,\text{\AA}^3
\qquad\text{and}\qquad
d_{\min}(\mathcal C)\ge 0.5\,\text{\AA},
\end{equation}
where \(d_{\min}(\mathcal C)\) is the minimum non-self interatomic distance in the constructed periodic structure.

Thus, the familiar condition \(d_{\min}\ge 0.5\,\text{\AA}\) together with \(V\ge 0.1\,\text{\AA}^3\) is the final structural-validity gate, but malformed samples may already be rejected earlier by wrapper- or construction-stage checks.

Let \(\mathcal G_{\mathrm{comp}}\), \(\mathcal G_{\mathrm{struct}}\), and \(\mathcal G_{\mathrm{val}}\) denote the subsets of generated crystals that pass the composition check, the structure check, and both checks, respectively. We then report
\begin{equation}
\mathrm{CompVal}=\frac{|\mathcal G_{\mathrm{comp}}|}{N_{\mathrm{gen}}},
\qquad
\mathrm{StructVal}=\frac{|\mathcal G_{\mathrm{struct}}|}{N_{\mathrm{gen}}},
\qquad
\mathrm{Val}=\frac{|\mathcal G_{\mathrm{val}}|}{N_{\mathrm{gen}}}.
\end{equation}
These validity metrics are reported for interpretability, but they are not the eligibility filter used for the main uniqueness, novelty, and \(\mathrm{UN}\) metrics.

\paragraph{Uniqueness, novelty, and \(\mathrm{UN}\).}
For the main DNG metrics, we first construct filtered generated and reference sets,
\[
\mathcal G_{\mathrm{eval}} \subseteq \mathcal G,
\qquad
\mathcal T_{\mathrm{eval}} \subseteq \mathcal T,
\]
by retaining only structures with finite geometry that satisfy the implemented \(N\)-ary threshold. In the current DNG code path, this threshold is \(\texttt{minimum\_nary}=1\), so unary structures are retained.

Structure comparisons are performed with \texttt{pymatgen}'s \texttt{StructureMatcher} using
\[
\mathrm{stol}=0.5,\qquad
\mathrm{ltol}=0.3,\qquad
\mathrm{angle\_tol}=10.
\]
A pair of structures is treated as matching whenever the matcher returns a valid alignment.

Let
\[
N_{\mathrm{eval}} = |\mathcal G_{\mathrm{eval}}|
\]
denote the number of generated structures that enter this evaluation stage. Uniqueness is computed by greedily deduplicating \(\mathcal G_{\mathrm{eval}}\), keeping only the first representative of each duplicate cluster. If \(N_{\mathrm{unique}}\) denotes the number of retained representatives, then
\begin{equation}
\mathrm{Unique}=\frac{N_{\mathrm{unique}}}{N_{\mathrm{eval}}}.
\end{equation}

Novelty is evaluated relative to the filtered reference set \(\mathcal T_{\mathrm{eval}}\), after the usual chemistry-system filtering used by the benchmark. Let \(N_{\mathrm{novel\_cand}}\) denote the number of generated structures that enter this novelty comparison, and let \(N_{\mathrm{novel}}\) denote the number of these structures that do not match any structure in \(\mathcal T_{\mathrm{eval}}\). We report
\begin{equation}
\mathrm{Novel}=\frac{N_{\mathrm{novel}}}{N_{\mathrm{novel\_cand}}}.
\end{equation}

The unique-and-novel set is not obtained by intersecting separately computed uniqueness and novelty flags. Instead, the code first restricts to the novel subset and then greedily deduplicates within that subset using the same first-occurrence rule as above. If \(N_{\mathrm{UN}}\) denotes the number of resulting representatives, then
\begin{equation}
\mathrm{UN}=\frac{N_{\mathrm{UN}}}{N_{\mathrm{novel\_cand}}}.
\end{equation}

In the usual non-degenerate case, \(N_{\mathrm{novel\_cand}}=N_{\mathrm{eval}}\), but we keep the notation separate here to reflect the implementation more faithfully.
\paragraph{Distribution matching.}
Distribution metrics are computed on the validity-filtered generated set \(\mathcal G_{\mathrm{val}}\). For any scalar crystal statistic \(x(\mathcal C)\), let \(P_x^{\mathrm{gen}}\) and \(P_x^{\mathrm{ref}}\) denote its empirical distributions over the generated and reference sets, respectively. We compare these distributions using the one-dimensional Wasserstein-1 distance
\begin{equation}
W_1(P,Q)=\int_{\mathbb R}\big|F_P(t)-F_Q(t)\big|\,dt,
\end{equation}
where \(F_P\) and \(F_Q\) are the corresponding cumulative distribution functions.

In the main text we report two such metrics. The first is based on mass density,
\begin{equation}
\rho(\mathcal C)=\frac{\mathrm{mass}(\mathcal C)}{\mathrm{vol}(\mathcal C)},
\end{equation}
and the second is based on the \(N\)-ary statistic,
\begin{equation}
n_{\mathrm{ary}}(\mathcal C)
=
\big|\{\text{elements present in }\mathcal C\}\big|.
\end{equation}
We therefore report
\begin{equation}
\mathrm{wdist}\text{-}\rho
=
W_1\!\big(P_{\rho}^{\mathrm{gen}},P_{\rho}^{\mathrm{ref}}\big),
\qquad
\mathrm{wdist}\text{-}N\text{-}\mathrm{ary}
=
W_1\!\big(P_{n_{\mathrm{ary}}}^{\mathrm{gen}},P_{n_{\mathrm{ary}}}^{\mathrm{ref}}\big).
\end{equation}

\paragraph{Thermodynamic stabilities.}
For offline evaluation, we generate \(10{,}000\) crystals and perform thermodynamic post-processing on all \(10{,}000\) generated structures. During training, we use a lighter version of this procedure, in which thermodynamic evaluation may be restricted to a smaller subset for efficiency.

Relaxation is performed with a compiled NequIP model using the batched TorchSim backend on CUDA, together with FIRE optimization and a Fr\'echet cell filter, so that both atomic positions and lattice degrees of freedom are optimized jointly. In this batched code path, relaxation is run for a fixed \(200\) FIRE steps; no force-threshold early stopping is used.

After relaxation, the implementation does not compute energy above hull via a hand-written subtraction formula. Instead, for each relaxed crystal \(\widetilde{\mathcal C}\) with final MLIP-predicted total energy \(E^{\mathrm{MLIP}}(\widetilde{\mathcal C})\), the code constructs a \texttt{ComputedStructureEntry}, attaches synthetic VASP-style metadata needed by \texttt{MaterialsProject2020Compatibility}, applies
\[
\texttt{MaterialsProject2020Compatibility(check\_potcar=False)},
\]
and then evaluates the corrected entry against the patched Materials Project phase diagram through \texttt{get\_e\_above\_hull(...)}. The reported quantity is therefore the hull distance of the corrected entry produced by this compatibility-processing pipeline.

Equivalently, one may view this as applying an MP2020-style correction to the relaxed MLIP energy before evaluating the distance to the reference convex hull, but the literal implementation is entry-based rather than an explicit subtraction against a separately written \(E_{\mathrm{hull}}^{\mathrm{ref}}\) term. If compatibility processing fails, returns no corrected entry, or produces a non-finite hull distance, the sample is recorded as a thermodynamic failure.

Internally, the thermo logger records two thresholds:
\begin{equation}
\mathrm{Stable}
=
\frac{1}{N_{\mathrm{thermo}}}
\big|\{\widetilde{\mathcal C}: e_{\mathrm{hull}}(\widetilde{\mathcal C})\le 0.0~\mathrm{eV/atom}\}\big|,
\end{equation}
and
\begin{equation}
\mathrm{Meta}
=
\frac{1}{N_{\mathrm{thermo}}}
\big|\{\widetilde{\mathcal C}: e_{\mathrm{hull}}(\widetilde{\mathcal C})\le 0.1~\mathrm{eV/atom}\}\big|,
\end{equation}
where \(N_{\mathrm{thermo}}\) is the number of crystals submitted to the thermodynamic pipeline. Relaxation and thermodynamic-processing failures count against these rates.

Thus, the implementation logs \(0.0\) eV/atom as \emph{stable} and \(0.1\) eV/atom as \emph{metastable}. In the main results, however, we often follow the common convention that the \(0.1\) eV/atom threshold is referred to simply as \emph{stable}. The appendix keeps the stricter logger terminology to match the implementation more closely.

Finally, we combine thermodynamic competitiveness with the unique-and-novel rate. Let \(\mathrm{Stable}_{\mathrm{UN}}\) and \(\mathrm{Meta}_{\mathrm{UN}}\) denote the fractions of unique-and-novel structures that satisfy the \(0.0\) and \(0.1\) eV/atom thresholds, respectively. We then define
\begin{equation}
\mathrm{SUN}
=
\mathrm{UN}\times \mathrm{Stable}_{\mathrm{UN}},
\qquad
\mathrm{MSUN}
=
\mathrm{UN}\times \mathrm{Meta}_{\mathrm{UN}}.
\end{equation}
Accordingly, when the main text informally treats the \(0.1\) eV/atom threshold as stability, it is this latter quantity that is being referred to.

\subsection{Crystal structure prediction (CSP)}
\label{apdx:csp_metrics}

Crystal structure prediction is a conditional task. For each test composition, the model generates a crystal conditioned on that composition, and the prediction is compared with the corresponding ground-truth structure \(\mathcal C_i^{\mathrm{gt}}\) using \texttt{pymatgen}'s \texttt{StructureMatcher}. Unless noted otherwise, we use the same matcher tolerances as in the DNG evaluation:
\[
\mathrm{stol}=0.5,\qquad
\mathrm{ltol}=0.3,\qquad
\mathrm{angle\_tol}=10.
\]

A prediction \(\widehat{\mathcal C}_i\) is counted as correct if \texttt{StructureMatcher} finds a valid match to \(\mathcal C_i^{\mathrm{gt}}\) under these tolerances. The match rate is therefore
\begin{equation}
\mathrm{MR}
=
\frac{1}{N_{\mathrm{test}}}
\big|
\{i : \widehat{\mathcal C}_i \text{ matches } \mathcal C_i^{\mathrm{gt}}\}
\big|,
\end{equation}
where \(N_{\mathrm{test}}\) is the number of test compositions.

For matched pairs, we additionally report the RMS displacement returned by the matcher after alignment. Let
\[
\mathcal M
=
\{i : \widehat{\mathcal C}_i \text{ matches } \mathcal C_i^{\mathrm{gt}}\}
\]
denote the set of matched test cases, and let \(r_i\) be the corresponding matcher RMS displacement for pair \(i\). We report
\begin{equation}
\mathrm{RMSD}
=
\frac{1}{|\mathcal M|}
\sum_{i\in\mathcal M} r_i.
\end{equation}

All CSP results in the main text use this standard single-sample setting.

\subsection{Sample-size intensive and extensive metrics}
\label{apdx:intensive_extensive_metrics}

An important practical point in \textit{de novo} generation is that not all metrics behave the same way when the number of generated samples changes. Some metrics describe the quality of a \emph{typical generated crystal}. Others describe the discovery yield of the \emph{entire generated set}. We refer to these two cases, by analogy with physics, as \emph{sample-intensive} and \emph{sample-extensive} metrics.

\paragraph{Sample-intensive metrics.}
A metric is sample-intensive if its target does not depend strongly on the total generation budget \(n\). These are quantities that can be estimated from a random subset of generated crystals without changing their meaning. In our setting, this includes:
\begin{itemize}
    \item compositional validity and structural validity,
    \item per-sample stability rates,
    \item average hull distance or other per-sample property means,
    \item distribution metrics such as Wasserstein distances on density or \(N\)-ary statistics.
\end{itemize}
For such quantities, a random subset gives an approximation to the same underlying target. In the simplest case, if \(g(\mathcal C_i)\) is a per-sample score or indicator, then
\begin{equation}
\widehat{\mu}_m = \frac{1}{m}\sum_{i=1}^m g(\mathcal C_i)
\end{equation}
is the natural estimator from a subset of size \(m\).

\paragraph{Sample-extensive metrics.}
A metric is sample-extensive if it depends directly on how many samples were generated. In crystal generation, this happens whenever duplicates matter. As the generation budget grows, duplicate collisions become more common, so the same model can look more or less diverse depending only on how many structures were sampled. In our setting, this includes:
\begin{itemize}
    \item uniqueness,
    \item the number of distinct discovered structures,
    \item novelty when reported as a discovery yield over the generated set,
    \item \(\mathrm{UN}\),
    \item \(\mathrm{SUN}\).
\end{itemize}
For example, if \(N_{\mathrm{unique}}(n)\) is the number of unique generated structures after drawing \(n\) samples, then
\begin{equation}
\mathrm{Unique}_n = \frac{N_{\mathrm{unique}}(n)}{n},
\qquad
\mathrm{UN}_n = \frac{N_{\mathrm{UN}}(n)}{n}
\end{equation}
are explicitly functions of \(n\). Evaluating these quantities on a smaller subset does \emph{not} estimate their value at the full budget. It simply computes the same metric at a different budget. In practice, this usually makes uniqueness and related discovery metrics look artificially better on small subsets.

This distinction explains why some metrics can be estimated on subsets and others cannot. Validity, stability, and average property metrics can be approximated on random subsets. By contrast, uniqueness, \(\mathrm{UN}\), and \(\mathrm{SUN}\) should be reported together with the number of generated samples and compared only at matched sample budgets.

A small caveat is that novelty can be defined in two different ways. If novelty is tested \emph{per sample} against a fixed reference set, then it behaves like an intensive quantity. In our setting, however, novelty is used as part of the deduplicated discovery pipeline, so it is more natural to treat it together with \(\mathrm{UN}\) and \(\mathrm{SUN}\) as a sample-extensive quantity.

\paragraph{Implication for SUN.}
This viewpoint also clarifies why it is reasonable to compute \(\mathrm{UN}\) on the full generated batch, but estimate stability only on a subset of the \(\mathrm{UN}\) structures. If \(\widehat p(\mathrm{stable}\mid \mathrm{UN})\) denotes the estimated stable fraction within the \(\mathrm{UN}\) set, then the natural estimator is
\begin{equation}
\widehat{\mathrm{SUN}}_n
=
\mathrm{UN}_n \times \widehat p(\mathrm{stable}\mid \mathrm{UN}).
\end{equation}
Here the first factor is a full-batch discovery statistic, while the second factor is a subset-based estimate of thermodynamic quality inside that discovered set.

\paragraph{Practical recommendation.}
For DNG evaluation, sample-extensive metrics such as uniqueness, \(\mathrm{UN}\), and \(\mathrm{SUN}\) should always be reported together with the generation budget \(n\). Sample-intensive metrics, such as validity, stability, and Wasserstein distances, can be estimated from random subsets when needed. This makes it easier to separate two different questions: whether the model generates \emph{good individual crystals}, and whether it continues to produce \emph{many distinct discoveries} as sampling is scaled up.

\clearpage
\section{Additional Results}
\label{apdx:add_results}
\subsection{Ablation of Subatomic Tokenization}
\label{apdx:ablation_subatomic}

To isolate the impact of Subatomic Tokenization, we train an identical Crystalite model using standard one-hot element encodings in place of PCA-compressed Subatomic Tokenization. Because the two embedding types have different dimensionalities, we balance the atom-type loss weight for the one-hot baseline so that its stability curve aligns with the subatomic-token model. Once the stability trajectories are matched, we compare the resulting unique-and-novel (UN) rate and the combined SUN score to evaluate the stability-diversity trade-off.

\begin{figure}[H]
    \centering
    \includegraphics[width=\linewidth]{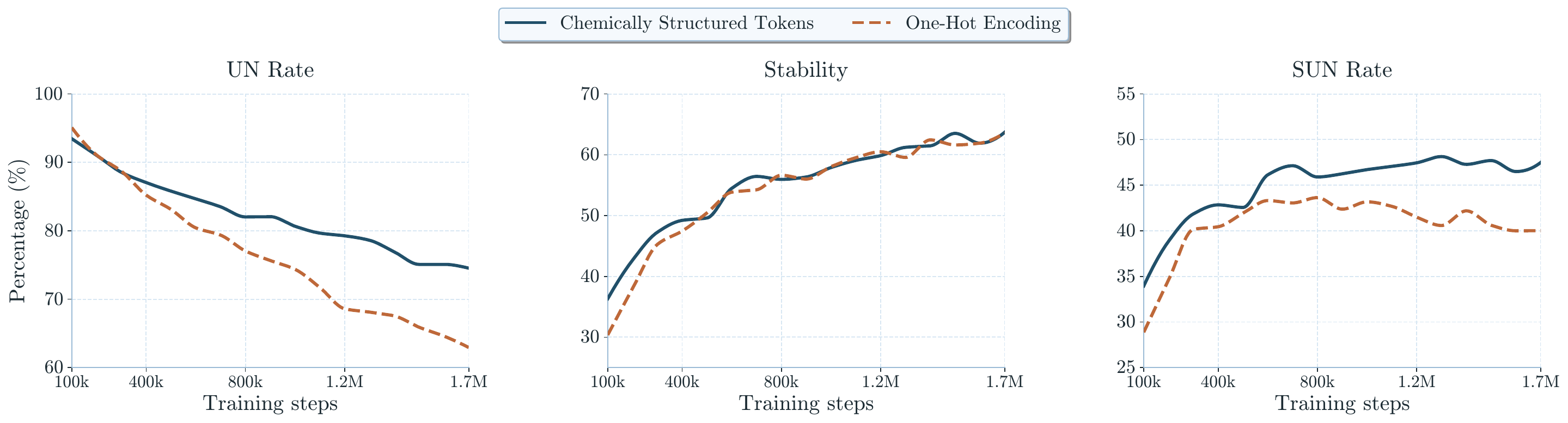}
    \caption{\textbf{Effect of Subatomic Tokenization on training dynamics.} The evolution of UN rate (left), stability (middle), and SUN rate (right) is compared between models using PCA-compressed Subatomic Tokenization and standard one-hot encoding. After balancing the loss weights to align the stability curves, Subatomic Tokenization achieves consistently higher UN and SUN scores throughout the latter half of training.}
    \label{fig:subatomic_ablation}
\end{figure}

Figure~\ref{fig:subatomic_ablation} illustrates this comparison. While both models follow the expected trajectory of increasing stability at the cost of diversity as training progresses, the model equipped with Subatomic Tokenization consistently maintains a better balance. In particular, it achieves a higher overall stability ceiling and sustains a higher SUN rate in the later stages of training. This indicates that a continuous subatomic tokenization provides an effective inductive bias for diffusion over atomic species, translating directly into an advantage for generative discovery performance.

\subsection{DNG MatterGen evaluation pipeline results}
Here we present our DNG metrics when evaluated using Mattergen evaluation pipeline, so that we can compare different models against ours on a setup that is not designed by us.

\begin{table}[H]
    \centering
    \small
    \caption{Validity, uniqueness, novelty, stability, and relaxation metrics using the MatterGen \citep{zeniGenerativeModelInorganic2025} evaluation pipeline for MP-20.}
    \label{tab:mattergen_pipeline}
    \renewcommand{\arraystretch}{1.1}
    \setlength{\tabcolsep}{4pt}
    \begin{tabular}{@{} l *{4}{S[table-format=3.2]} S[table-format=2.2] S[table-format=2.2] S[table-format=1.3] S[table-format=1.3] @{}}
        \toprule
        \multirow{3}{*}{\textbf{Model}}
        & \multicolumn{4}{c}{\textbf{Validity and Novelty}}
        & \multicolumn{4}{c}{\textbf{Stability and Relaxation}} \\
        \cmidrule(lr){2-5} \cmidrule(lr){6-9}
        & {Struct. Val.}
        & {Comp. Val.}
        & {Unique}
        & {Novel}
        & {Stable}
        & {S.U.N.}
        & {Avg. Hull}
        & {Avg. RMSD} \\
        & {(\%) $\uparrow$}
        & {(\%) $\uparrow$}
        & {(\%) $\uparrow$}
        & {(\%) $\uparrow$}
        & {(\%) $\uparrow$}
        & {(\%) $\uparrow$}
        & {(eV/atom) $\downarrow$}
        & {(\AA) $\downarrow$} \\
        \midrule
        MatterGen & {\bfseries 100.00} & 83.48 & {\bfseries 97.94} & {\bfseries 75.02} & 45.74 & 23.75 & 0.182 & {\bfseries 0.153} \\
        ADiT & {\bfseries 100.00} & {\bfseries 90.24} & 89.62 & 43.15 & {\bfseries 69.96} & 17.17 &  0.148 & 0.493 \\
        \midrule
        \textbf{Crystalite} & {\bfseries 100.00} & 84.62 & 94.73 & 56.63 & { 64.52} & {\bfseries 24.26} & {\bfseries 0.145} & 0.274 \\
        \bottomrule
    \end{tabular}
\end{table}

\subsection{GEM effect on DNG Results}\label{apdx:GEM_DNG}

\begin{figure}[H]
    \centering
    \includegraphics[width=1.0\linewidth]{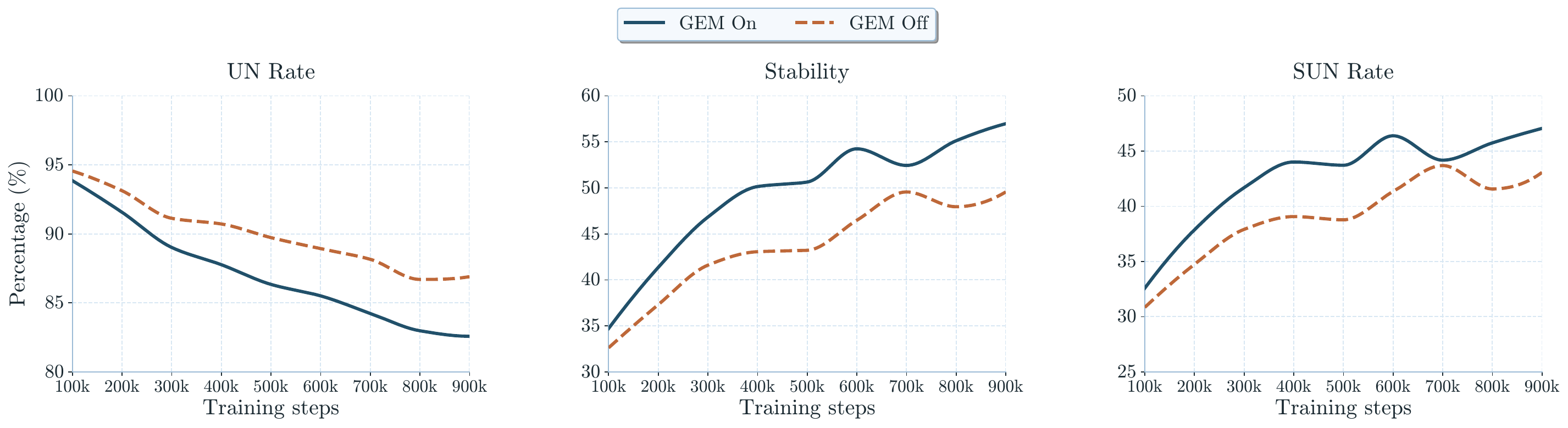}
    \caption{\textbf{Effect of GEM on \textit{de novo} generation training dynamics.} UN rate (left), stability (middle), and SUN rate (right) as a function of training steps, with and without GEM.}
    \label{fig:GEM_dng}
\end{figure}

Figure~\ref{fig:GEM_dng} compares the training dynamics of Crystalite with and without the Geometry Enhancement Module (GEM) in the \textit{de novo} generation setting. Both configurations exhibit the expected decline in the unique-and-novel (UN) rate as training progresses, reflecting the general trade-off between diversity and stability. However, the model with GEM learns substantially faster on the stability axis and maintains higher stability throughout training. As a result, it also achieves a consistently higher Stable, Unique, and Novel (SUN) rate across the full training trajectory. This suggests that injecting periodic pairwise geometry into attention improves the structural quality of generated crystals without causing a disproportionate loss in generative diversity.

\subsection{GEM effect on CSP Results}\label{apdx:GEM_CSP}

\begin{figure}[H]
    \centering
    \includegraphics[width=1.0\linewidth]{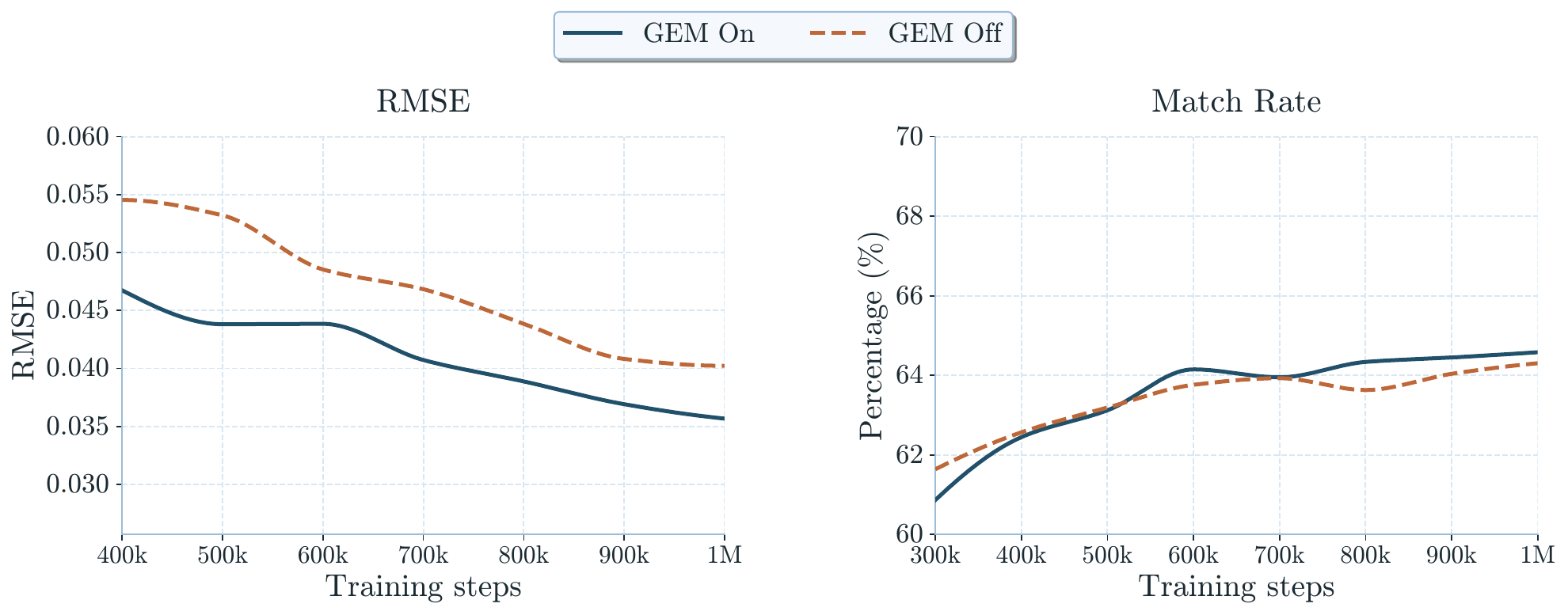}
    \caption{\textbf{Effect of GEM on crystal structure prediction.} RMSE (left) and Match Rate (right) as a function of training steps, with and without GEM.}
    \label{fig:GEM_csp}
\end{figure}

Figure~\ref{fig:GEM_csp} shows the corresponding ablation for crystal structure prediction (CSP). Here, GEM has only a modest effect on Match Rate (MR), but leads to a clearer and more consistent improvement in RMSE throughout training. In other words, GEM appears to have a limited effect on whether the model recovers the correct structural mode, but a stronger effect on how accurately that structure is refined once recovered. This is consistent with the interpretation that the geometric biases introduced by GEM primarily improve local atomic placement and overall structural fidelity during denoising.

\subsection{DNG Sensitivity to anti-annealing}
\label{apdx:dng_anti_annealing}
We also ablate the channel-wise anti-annealing settings used at sampling time, varying the strength of anti-annealing for the coordinate and lattice channels while keeping the trained model fixed. 

\begin{table}[H]
    \centering
    \scriptsize
    \caption{Generative quality, diversity, stability, and distribution metrics for Crystalite across the aa grid.}
    \label{tab:crystalite_aa_grid}
    \renewcommand{\arraystretch}{1.1}
    \setlength{\tabcolsep}{4pt}
    \resizebox{\linewidth}{!}{%
    \begin{tabular}{@{} S[table-format=2.0] S[table-format=2.0] S[table-format=2.0] *{5}{S[table-format=3.2]} S[table-format=2.2] S[table-format=2.2] S[table-format=1.3] S[table-format=1.3] @{}}
        \toprule
        \multicolumn{3}{c}{\textbf{AA settings}}
        & \multicolumn{5}{c}{\textbf{Quality and Diversity}}
        & \multicolumn{4}{c}{\textbf{Stability and Distribution}} \\
        \cmidrule(lr){1-3} \cmidrule(lr){4-8} \cmidrule(lr){9-12}
        {$\mathrm{aa}_{\mathrm{coords}}$}
        & {$\mathrm{aa}_{\mathrm{types}}$}
        & {$\mathrm{aa}_{\mathrm{lattice}}$}
        & {Struct. Val.}
        & {Comp. Val.}
        & {Unique}
        & {Novel}
        & {U.N.}
        & {Stable}
        & {S.U.N.}
        & {wdist-$\rho$}
        & {wdist N-ary} \\
        & & &
        {(\%) $\uparrow$}
        & {(\%) $\uparrow$}
        & {(\%) $\uparrow$}
        & {(\%) $\uparrow$}
        & {(\%) $\uparrow$}
        & {(\%) $\uparrow$}
        & {(\%) $\uparrow$}
        & {$\downarrow$}
        & {$\downarrow$} \\
        \midrule
        0 & 0 & 0 & 99.76 & 81.49 & 98.78 & 86.04 & 85.55 & 63.28 & 49.07 & 0.125 & 0.200 \\
        0 & 0 & 4 & 99.71 & 83.30 & 98.58 & 83.15 & 82.37 & 66.75 & 49.37 & 0.428 & 0.221 \\
        0 & 0 & 10 & 99.71 & 81.25 & 99.02 & 86.28 & 85.94 & 62.74 & 48.93 & 0.131 & {\bfseries 0.191} \\
        4 & 0 & 0 & 99.76 & 80.91 & 99.22 & 86.18 & 85.99 & 61.18 & 47.22 & 0.179 & 0.249 \\
        4 & 0 & 4 & 99.85 & 81.69 & 98.44 & 83.94 & 83.20 & {\bfseries 68.12} & {\bfseries 51.42} & 0.500 & 0.226 \\
        4 & 0 & 10 & 99.71 & 80.57 & 99.02 & 85.94 & 85.55 & 62.21 & 47.90 & 0.176 & 0.248 \\
        10 & 0 & 0 & {\bfseries 99.90} & 81.59 & 98.83 & 86.47 & 86.04 & 62.89 & 49.12 & {\bfseries 0.111} & 0.205 \\
        10 & 0 & 4 & 99.66 & 83.15 & 98.58 & 82.91 & 82.13 & 66.80 & 49.12 & 0.421 & 0.205 \\
        10 & 0 & 10 & 99.76 & 80.81 & 98.88 & 85.79 & 85.40 & 63.62 & 49.22 & 0.125 & 0.198 \\
        \midrule
        0 & 10 & 0 & 99.76 & 81.49 & 98.93 & 86.43 & 85.99 & 62.06 & 48.29 & 0.117 & 0.199 \\
        0 & 10 & 4 & 99.80 & 82.91 & 98.54 & 83.20 & 82.47 & 67.04 & 49.66 & 0.401 & 0.210 \\
        0 & 10 & 10 & 99.66 & 80.91 & 98.97 & 86.33 & 85.89 & 62.65 & 48.78 & 0.126 & 0.196 \\
        4 & 10 & 0 & 99.71 & 80.03 & 99.27 & 86.38 & 86.13 & 61.04 & 47.27 & 0.168 & 0.247 \\
        4 & 10 & 4 & 99.76 & 81.93 & 98.63 & 83.84 & 83.25 & 67.33 & 50.73 & 0.482 & 0.231 \\
        4 & 10 & 10 & 99.80 & 79.88 & {\bfseries 99.32} & 85.94 & 85.74 & 60.64 & 46.48 & 0.155 & 0.228 \\
        10 & 10 & 0 & 99.85 & 81.15 & 98.93 & 86.43 & 86.04 & 63.13 & 49.41 & 0.123 & 0.212 \\
        10 & 10 & 4 & 99.71 & {\bfseries 83.35} & 98.54 & 82.81 & 82.03 & 67.04 & 49.27 & 0.416 & 0.209 \\
        10 & 10 & 10 & 99.85 & 81.79 & 98.93 & 86.28 & 85.84 & 62.55 & 48.63 & 0.125 & 0.210 \\
        \midrule
        0 & 20 & 0 & 99.76 & 81.15 & 98.93 & {\bfseries 86.62} & {\bfseries 86.18} & 62.35 & 48.78 & 0.131 & 0.214 \\
        0 & 20 & 4 & 99.80 & 82.96 & 98.44 & 82.86 & 81.98 & 66.80 & 48.97 & 0.415 & 0.215 \\
        0 & 20 & 10 & 99.85 & 81.69 & 98.97 & 86.38 & 85.94 & 63.04 & 49.22 & 0.120 & 0.197 \\
        4 & 20 & 0 & 99.85 & 80.91 & 99.17 & 86.18 & 85.89 & 60.94 & 46.92 & 0.180 & 0.240 \\
        4 & 20 & 4 & 99.85 & 82.13 & 98.39 & 83.94 & 83.11 & 67.58 & 50.78 & 0.477 & 0.226 \\
        4 & 20 & 10 & 99.76 & 80.27 & 99.07 & 85.84 & 85.50 & 60.94 & 46.48 & 0.187 & 0.247 \\
        10 & 20 & 0 & {\bfseries 99.90} & 81.20 & 98.93 & 86.33 & 85.89 & 62.89 & 49.02 & 0.133 & 0.210 \\
        10 & 20 & 4 & 99.66 & {\bfseries 83.35} & 98.63 & 83.15 & 82.47 & 67.04 & 49.71 & 0.412 & 0.220 \\
        10 & 20 & 10 & 99.85 & 81.64 & 98.88 & 86.38 & 85.99 & 62.89 & 49.12 & 0.130 & 0.202 \\
        \bottomrule
    \end{tabular}%
    }
\end{table}

Overall, the results are fairly insensitive to this choice: across a reasonable range of settings, the main conclusions remain unchanged and Crystalite performs consistently well. Although one anti-annealing configuration achieved the highest SUN score, it also produced noticeably worse Wasserstein distances, indicating poorer distributional alignment. For this reason, we do not report the single best-SUN configuration, but instead select a more balanced setting that preserves strong discovery performance while maintaining better agreement with the reference distribution. This suggests that anti-annealing is a useful but non-fragile sampling heuristic, and that the reported results do not depend critically on a finely tuned choice of anti-annealing parameters.

\subsection{CSP Sensitivity to anti-annealing}
\label{apdx:csp_anti_annealing}
We also ablate the channel-wise anti-annealing settings used at sampling time for the CSP model, varying the strength of anti-annealing for the coordinate and lattice channels while keeping the trained model fixed.

\begin{table}[H]
    \centering
    \scriptsize
    \caption{Match rate and RMSE metrics for varying anti-annealing settings on coordinate and lattice channels in CSP.}
    \label{tab:csp_aa_grid}
    \renewcommand{\arraystretch}{1.1}
    \setlength{\tabcolsep}{6pt}
    \begin{tabular}{@{} S[table-format=2.0] S[table-format=2.0] c c @{}}
        \toprule
        \multicolumn{2}{c}{\textbf{AA settings}}
        & \multicolumn{2}{c}{\textbf{Metrics}} \\
        \cmidrule(lr){1-2} \cmidrule(lr){3-4}
        {$\mathrm{aa}_{\mathrm{coords}}$}
        & {$\mathrm{aa}_{\mathrm{lattice}}$}
        & {Match Rate (\%) $\uparrow$}
        & {RMSE ($\times 10^{-2}$) $\downarrow$} \\
        \midrule
        0 & 0 & $65.04 \pm 0.23$ & $4.11 \pm 0.05$ \\
        0 & 4 & $65.22 \pm 0.16$ & $4.14 \pm 0.05$ \\
        0 & 10 & $65.01 \pm 0.22$ & $4.12 \pm 0.05$ \\
        0 & 20 & $65.02 \pm 0.22$ & $4.11 \pm 0.06$ \\
        \midrule
        4 & 0 & $65.96 \pm 0.17$ & $3.41 \pm 0.07$ \\
        4 & 4 & $\mathbf{66.09 \pm 0.09}$ & $\mathbf{3.37 \pm 0.05}$ \\
        4 & 10 & $65.94 \pm 0.15$ & $3.39 \pm 0.07$ \\
        4 & 20 & $65.96 \pm 0.16$ & $3.41 \pm 0.08$ \\
        \midrule
        10 & 0 & $65.15 \pm 0.19$ & $4.19 \pm 0.04$ \\
        10 & 4 & $65.22 \pm 0.08$ & $4.12 \pm 0.04$ \\
        10 & 10 & $65.13 \pm 0.14$ & $4.15 \pm 0.05$ \\
        10 & 20 & $65.17 \pm 0.17$ & $4.20 \pm 0.05$ \\
        \midrule
        20 & 0 & $65.19 \pm 0.15$ & $4.19 \pm 0.08$ \\
        20 & 4 & $65.26 \pm 0.11$ & $4.16 \pm 0.08$ \\
        20 & 10 & $65.17 \pm 0.16$ & $4.21 \pm 0.07$ \\
        20 & 20 & $65.13 \pm 0.14$ & $4.20 \pm 0.06$ \\
        \midrule
        27 & 36 & $65.13 \pm 0.00$ & $4.07 \pm 0.00$ \\
        \bottomrule
    \end{tabular}
\end{table}

Similar to the DNG results, CSP performance exhibits broad stability across a range of anti-annealing settings, confirming that the sampling process is generally robust. However, unlike the metric trade-offs observed in the DNG ablations, we find a clear and consistent alignment between the CSP metrics: the configuration with modest anti-annealing on both channels (\(\mathrm{aa}_{\mathrm{coords}}=4\), \(\mathrm{aa}_{\mathrm{lattice}}=4\)) simultaneously achieves the highest match rate and the lowest RMSE. This optimal pocket provides a distinct performance advantage without inducing instability. Therefore, we highlight this balanced configuration, demonstrating that while the model does not depend critically on highly tuned parameters to perform well, targeted anti-annealing still yields a measurable boost to structural prediction accuracy.

\clearpage
\section{Crystalite S.U.N. Crystals}
\label{apdx:crys}

\begin{figure}[htbp]
    \centering

    \begin{subfigure}{0.3\textwidth}
        \centering
        \includegraphics[width=\linewidth]{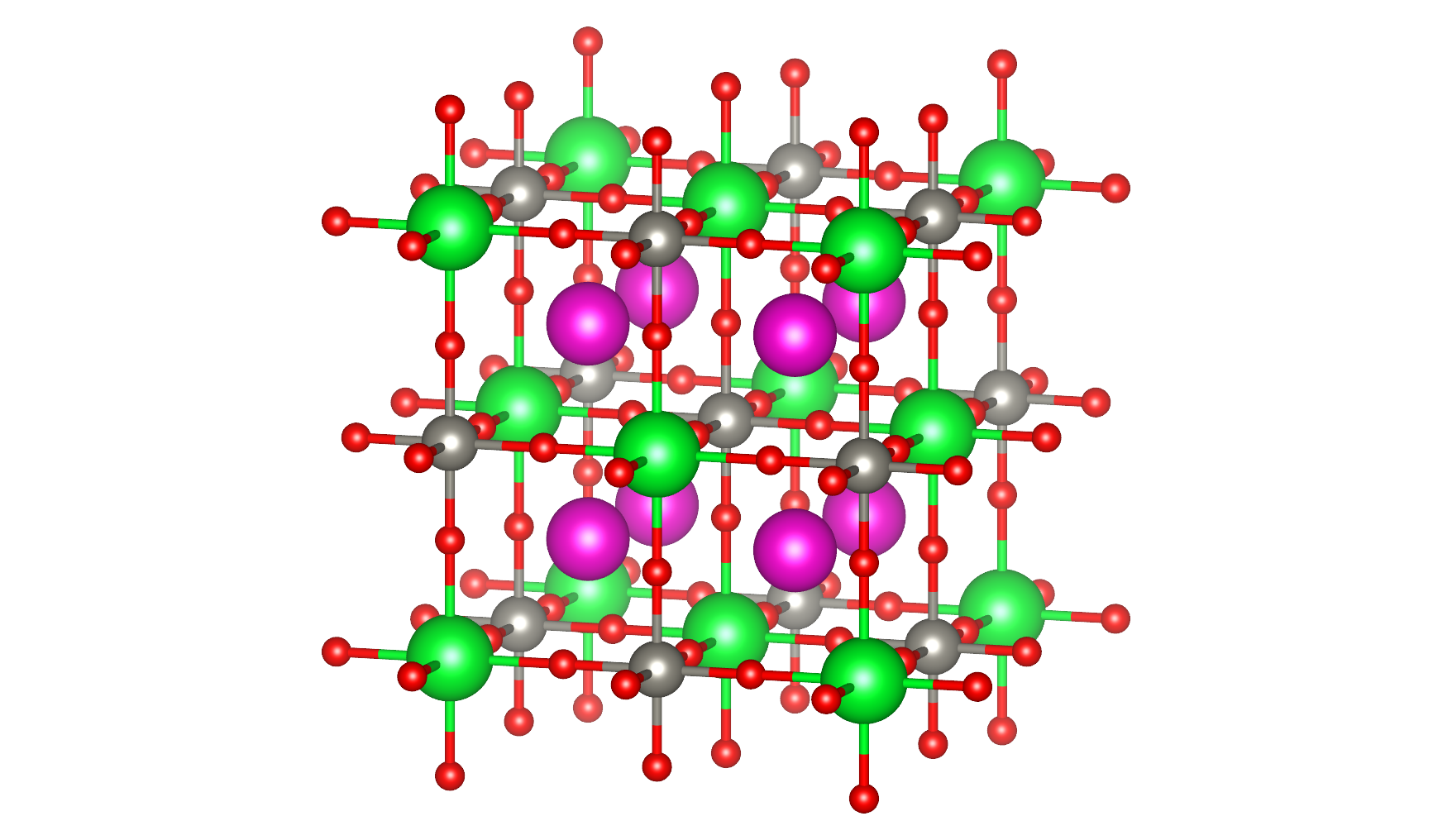}
        \caption{Sr${}_4$Eu${}_8$W$_4$O${}_{24}$}
    \end{subfigure}
    \begin{subfigure}{0.3\textwidth}
        \centering
        \includegraphics[width=\linewidth]{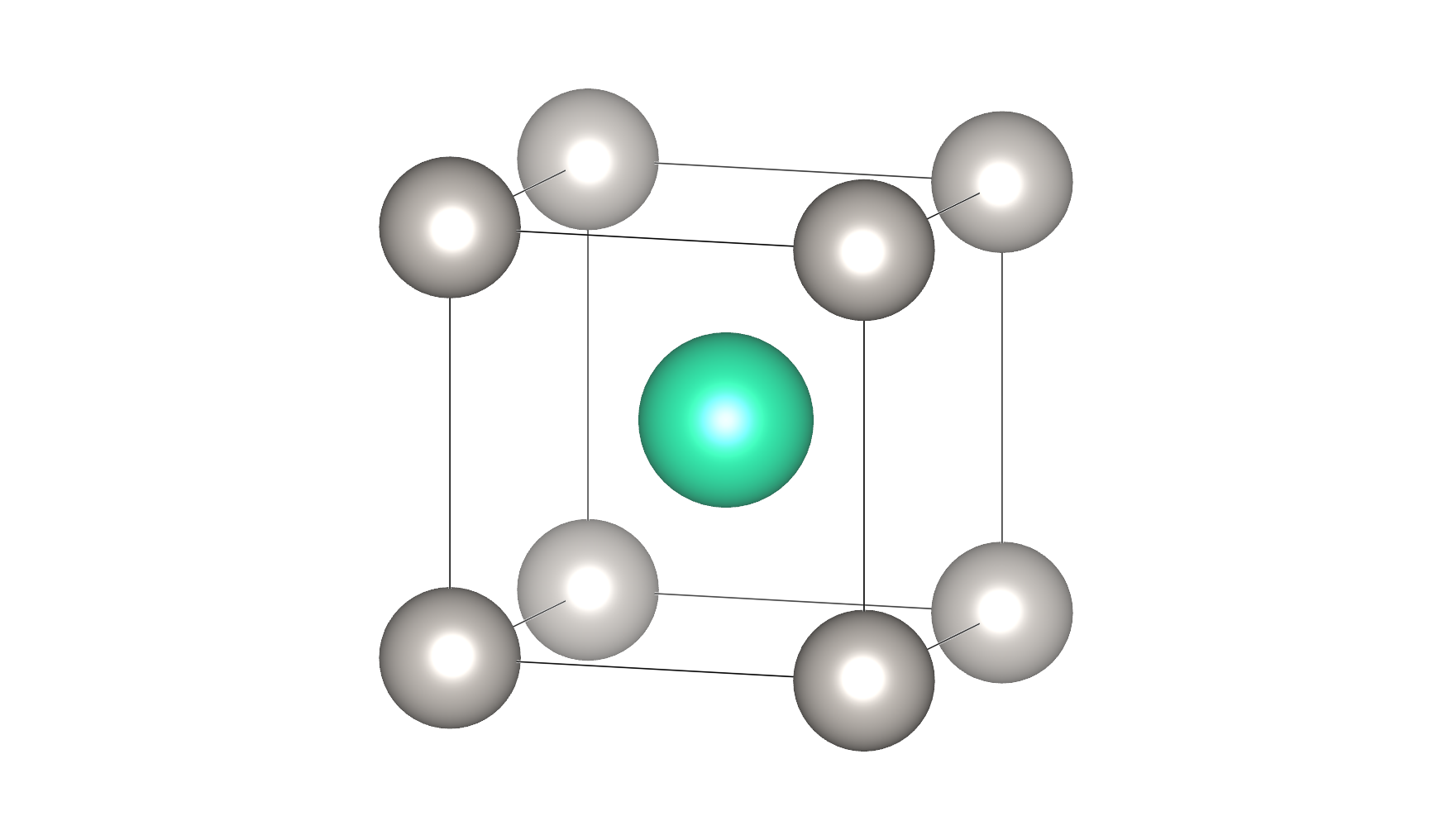}
        \caption{LuPt}
    \end{subfigure}
    \begin{subfigure}{0.3\textwidth}
        \centering
        \includegraphics[width=\linewidth]{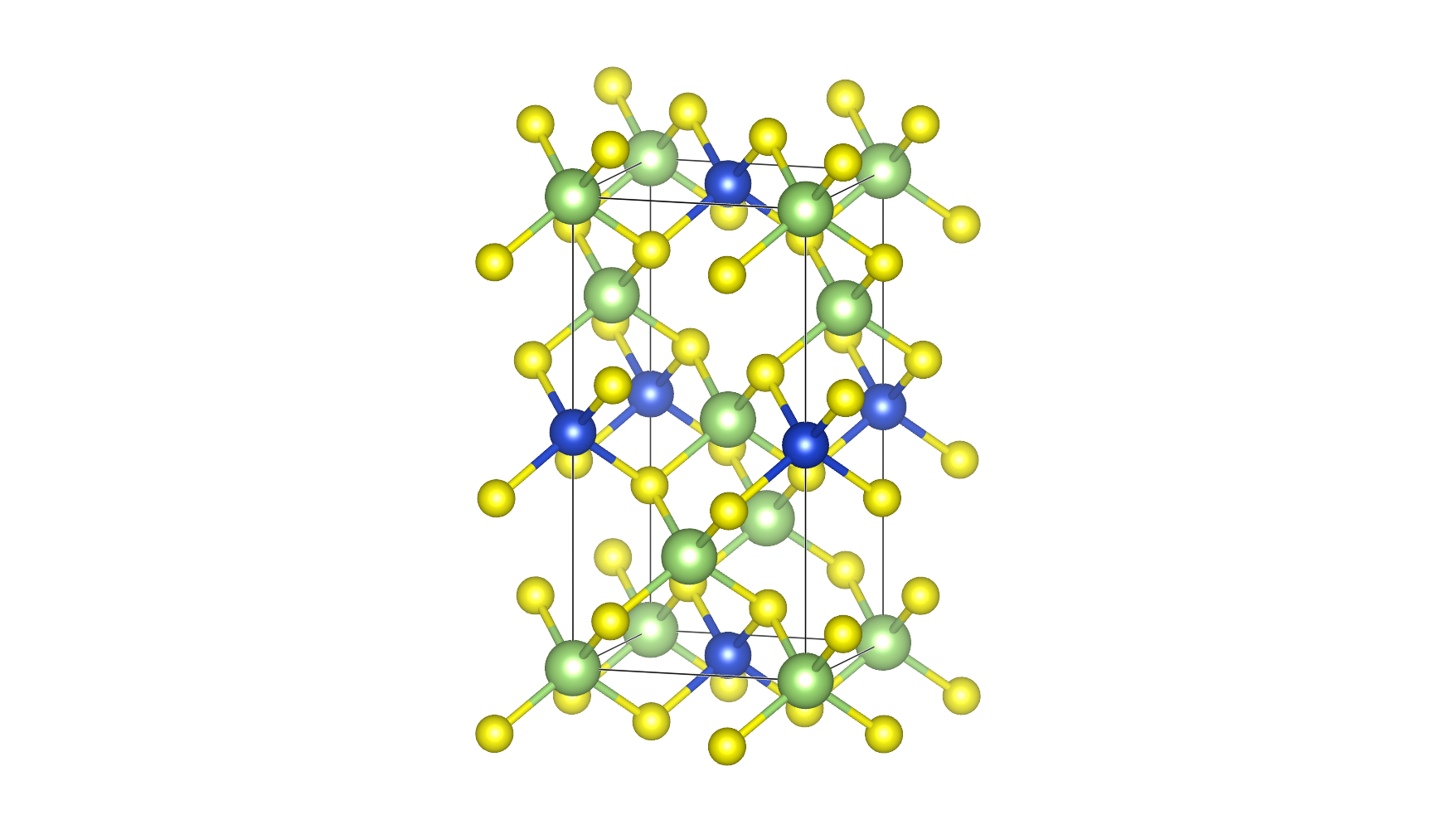}
        \caption{Ga${}_4$Cu${}_2$S${}_8$}
    \end{subfigure}

    \vspace{0.5em}

    \begin{subfigure}{0.3\textwidth}
        \centering
        \includegraphics[width=\linewidth]{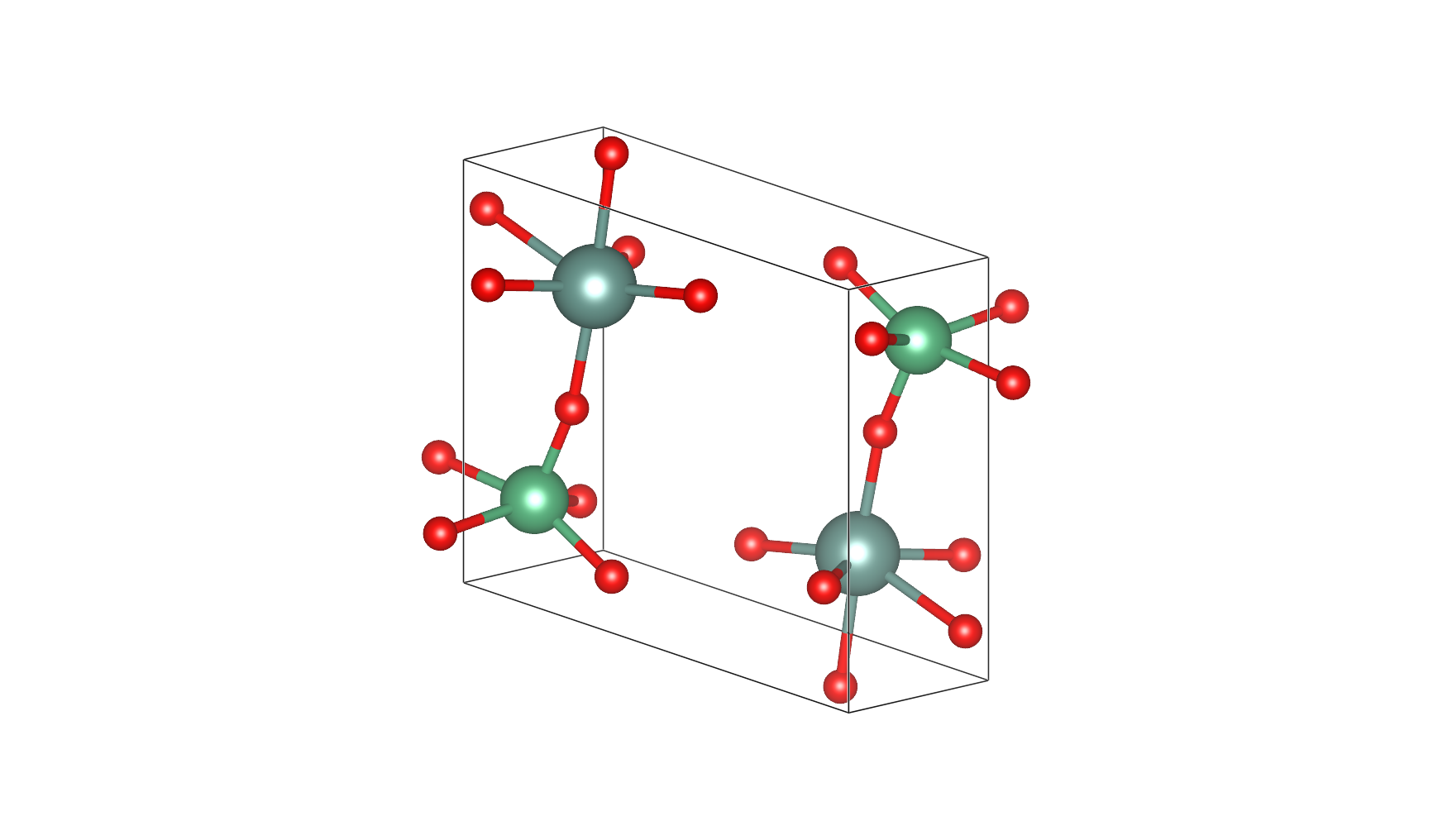}
        \caption{Y$_2$Nb$_2$O$_8$}
    \end{subfigure}
    \begin{subfigure}{0.3\textwidth}
        \centering
        \includegraphics[width=\linewidth]{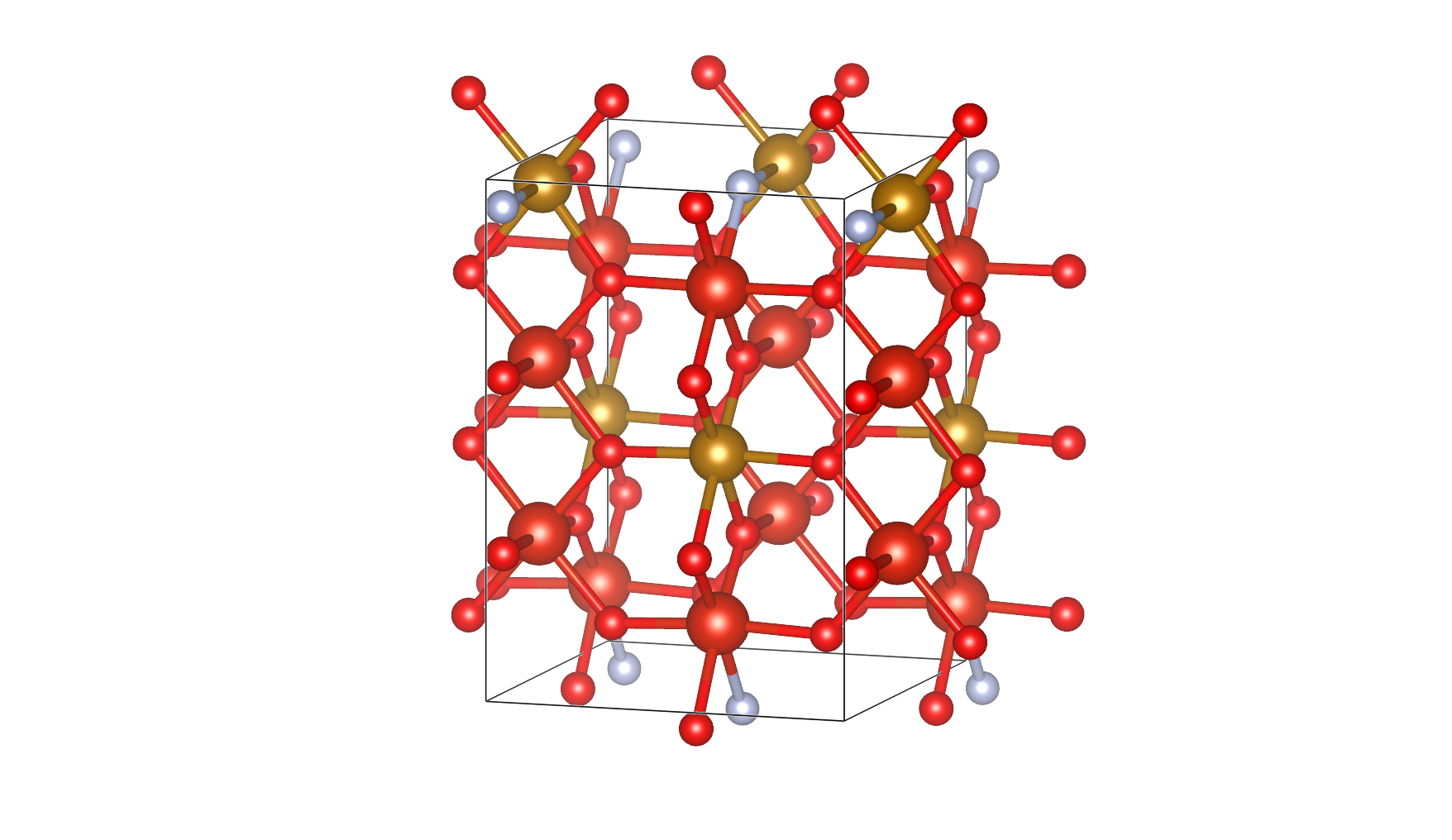}
        \caption{V$_8$Fe$_4$O$_{22}$F$_2$}
    \end{subfigure}
    \begin{subfigure}{0.3\textwidth}
        \centering
        \includegraphics[width=\linewidth]{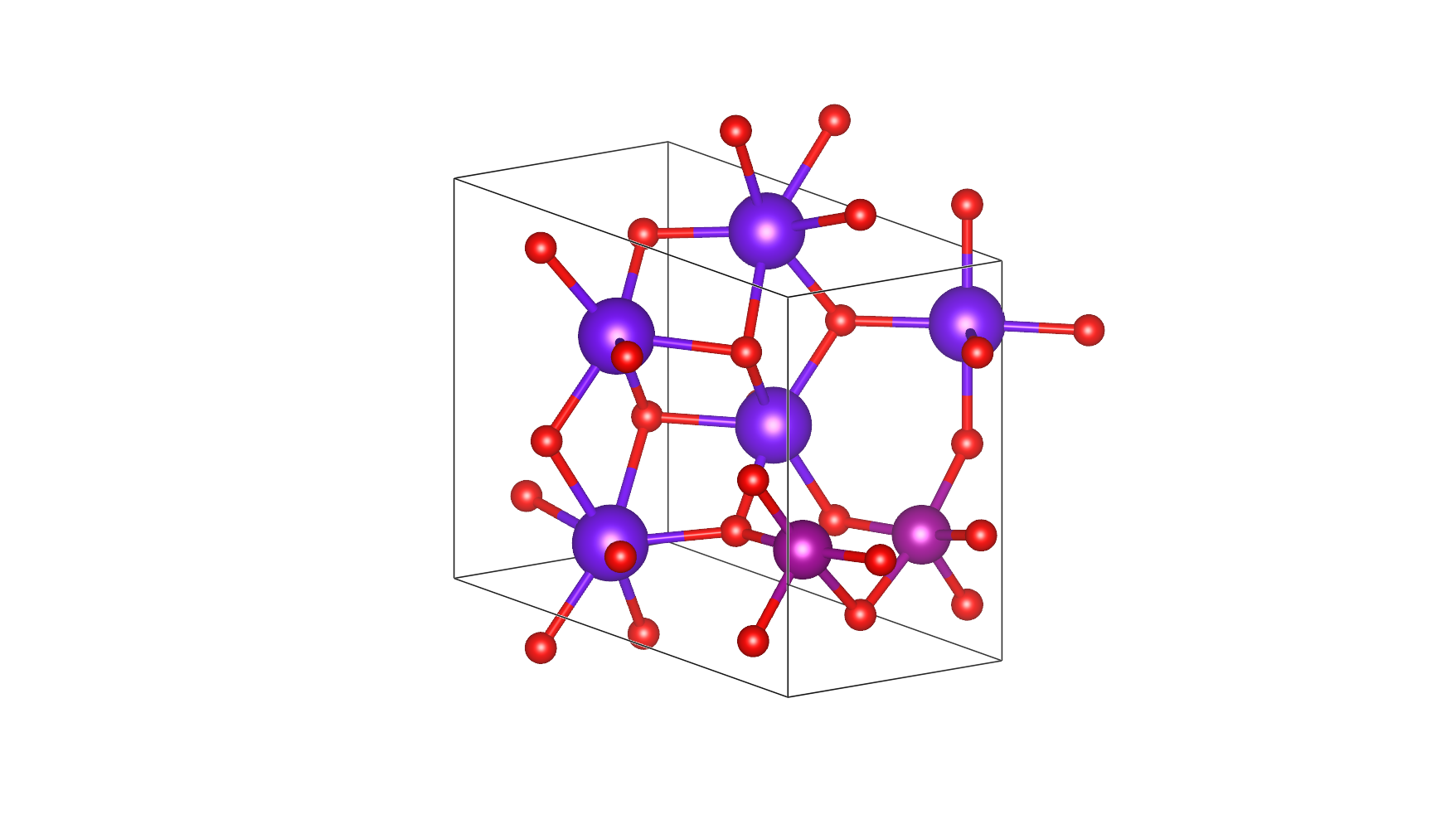}
        \caption{Tb$_5$Mn$_2$O$_{11}$}
    \end{subfigure}

    \vspace{0.5em}

    \begin{subfigure}{0.3\textwidth}
        \centering
        \includegraphics[width=\linewidth]{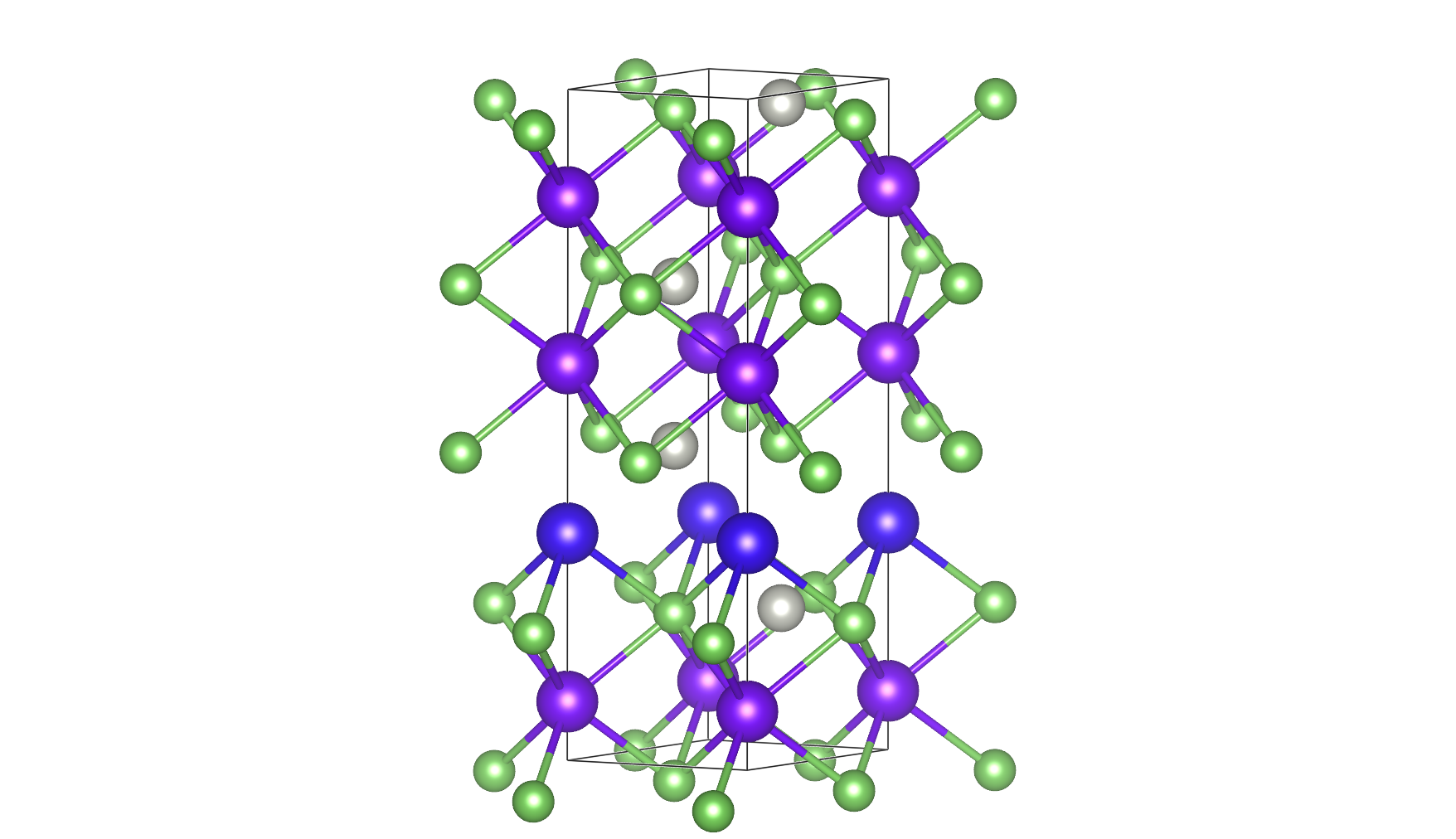}
        \caption{Tb$_3$DyAs$_4$Pd$_4$}
    \end{subfigure}
    \begin{subfigure}{0.3\textwidth}
        \centering
        \includegraphics[width=\linewidth]{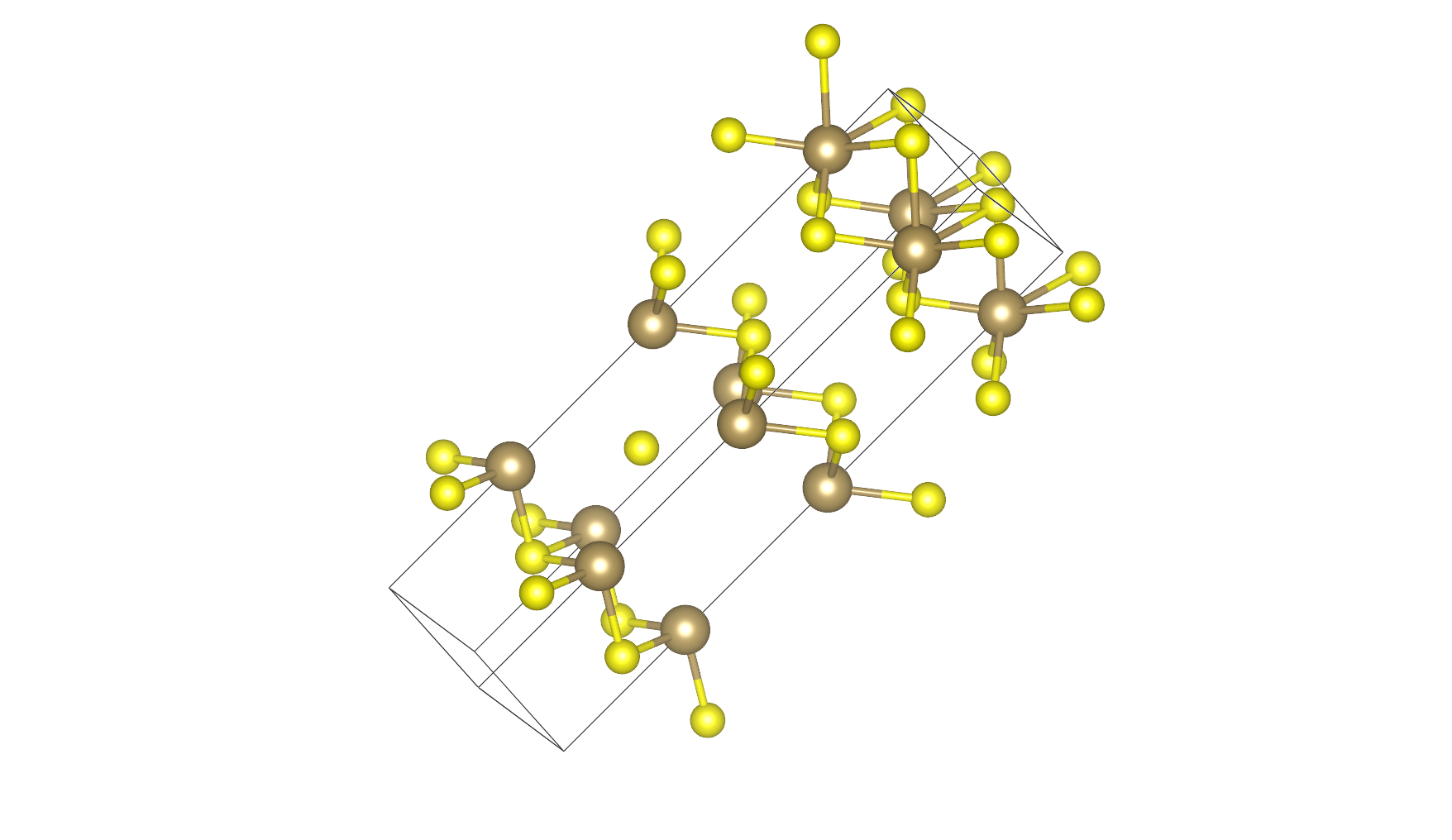}
        \caption{Ta$_3$S$_5$}
    \end{subfigure}
    \begin{subfigure}{0.3\textwidth}
        \centering
        \includegraphics[width=\linewidth]{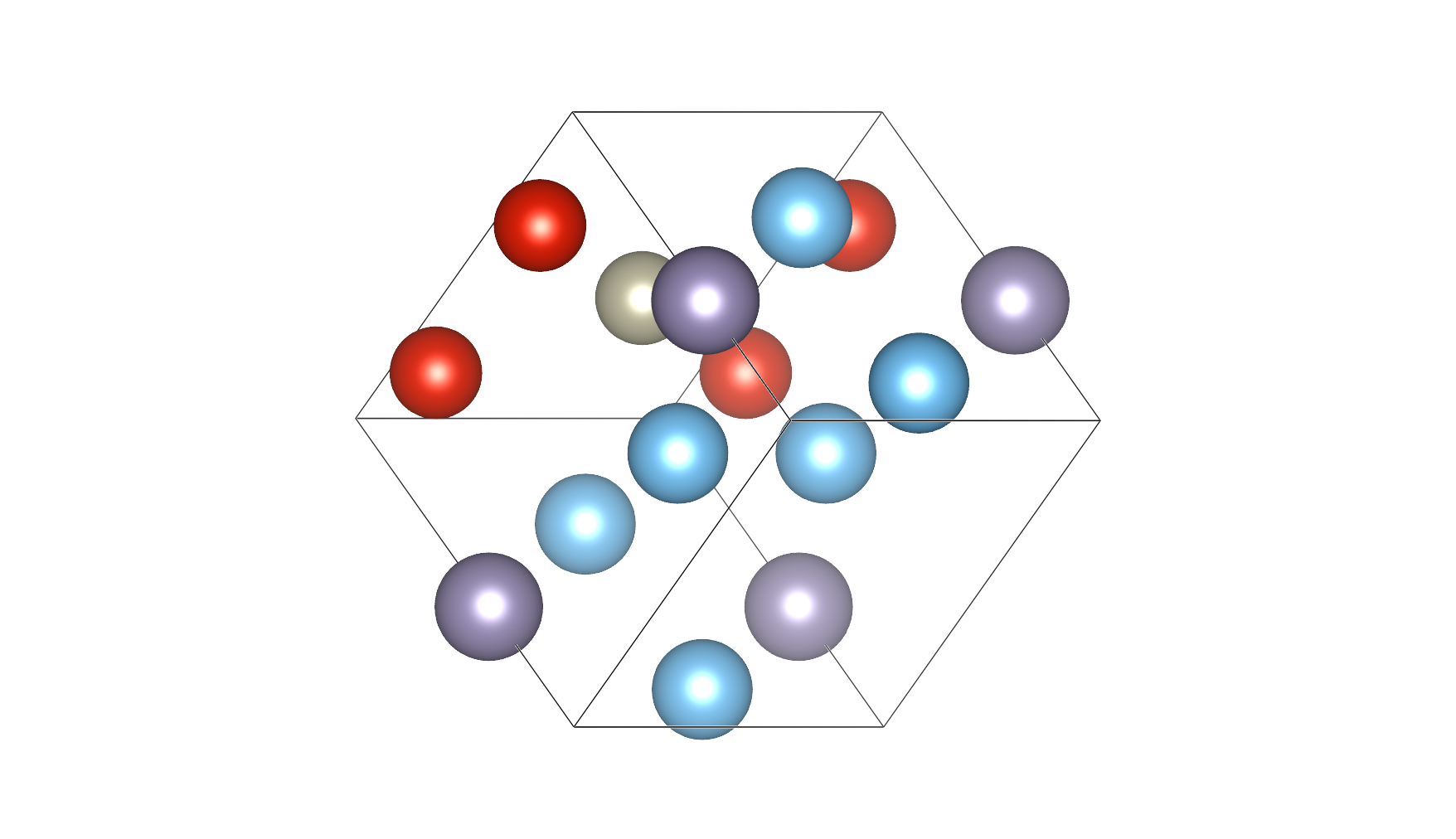}
        \caption{Ti$_4$V$_2$ReSn}
    \end{subfigure}

    \caption{A set of stable, unique and novel crystals generated by Crystalite trained on MP-20.}
    \label{fig:grid9}
\end{figure}

\end{document}